\documentclass[11pt]{article}

\usepackage{acl}

\usepackage{times}
\usepackage{latexsym}

\usepackage[T1]{fontenc}
\usepackage[utf8]{inputenc}

\usepackage{microtype}

\usepackage{inconsolata}

\usepackage{graphicx}
\usepackage{booktabs}
\usepackage{multirow}
\usepackage{tabularx}
\usepackage{makecell}
\usepackage{amsmath}
\usepackage{amssymb}
\usepackage{bm}
\usepackage{tcolorbox}
\usepackage{xcolor}

\title{\raisebox{-0.4\height}{\includegraphics[width=1.5cm]{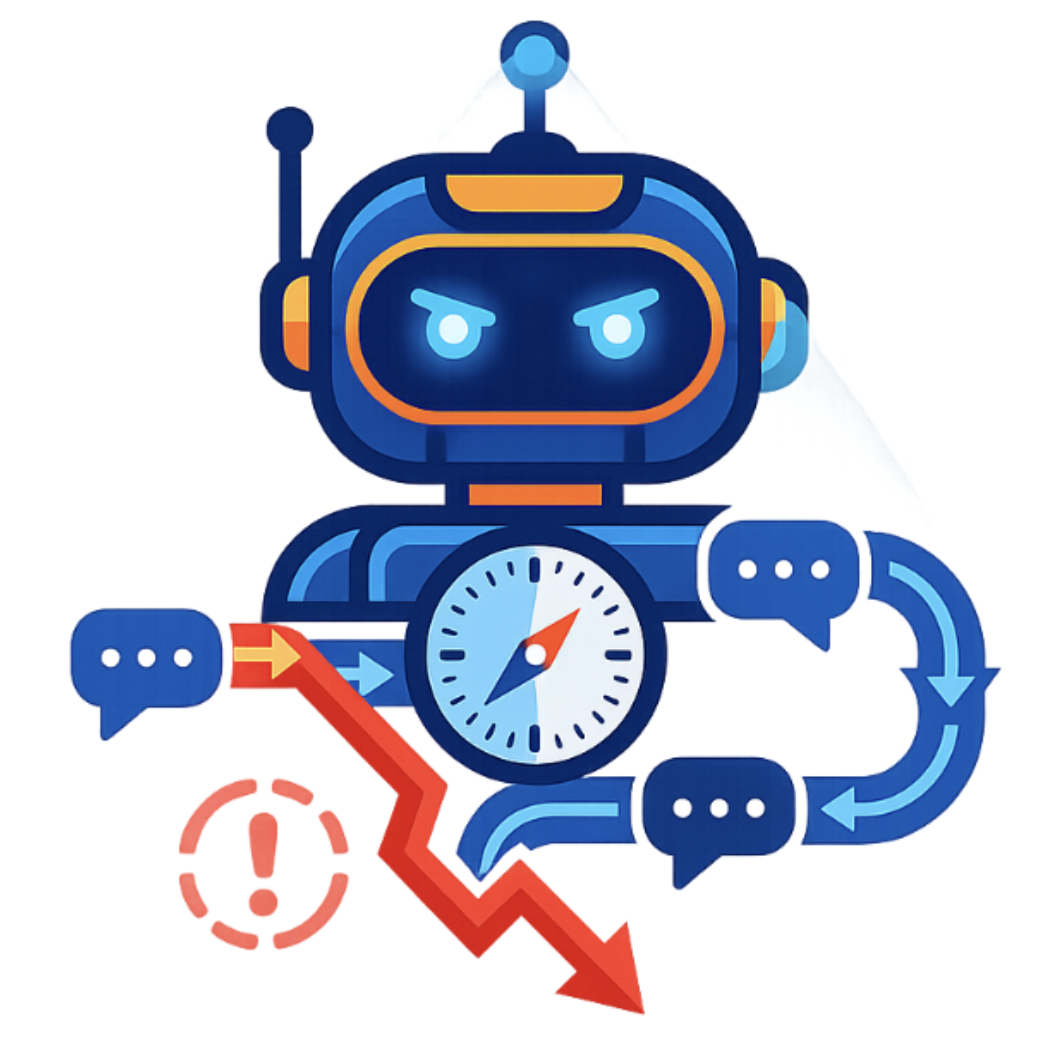}}\, TRACER: Early Failure Detection for Task-Oriented Dialogue}

\author{Erfan Nourbakhsh, Rocky Slavin, Ke Yang, \and Anthony Rios \\
  The University of Texas at San Antonio \\
  \texttt{\{erfan.nourbakhsh, anthony.rios\}@utsa.edu} \\}

\begin{document}
\maketitle
\begin{abstract}
Task-oriented dialogue systems often fail before the final breakdown is obvious, but most evaluation only measures failure after the conversation has already gone wrong. We present TRACER, a method for early failure detection in task-oriented dialogue. TRACER predicts from a partial dialogue whether the full conversation will eventually fail by combining simple trajectory signals from belief-state changes with text representations of the evolving dialogue state. We evaluate the method in both oracle and generated belief-state settings, and test how well it works when only 25\%, 50\%, 75\%, or 100\% of the dialogue is visible. Across these settings, TRACER detects useful failure signals well before the end of the conversation and outperforms heuristic, classical, and single-stream baselines. These results suggest that early failure detection can provide a practical warning signal for dialogue systems before the interaction fully breaks down. Source code can be found here: \url{https://github.com/erfan-nourbakhsh/TRACER}.
 \end{abstract}

\begin{figure}[t]
    \centering
    \includegraphics[width=0.9\linewidth]{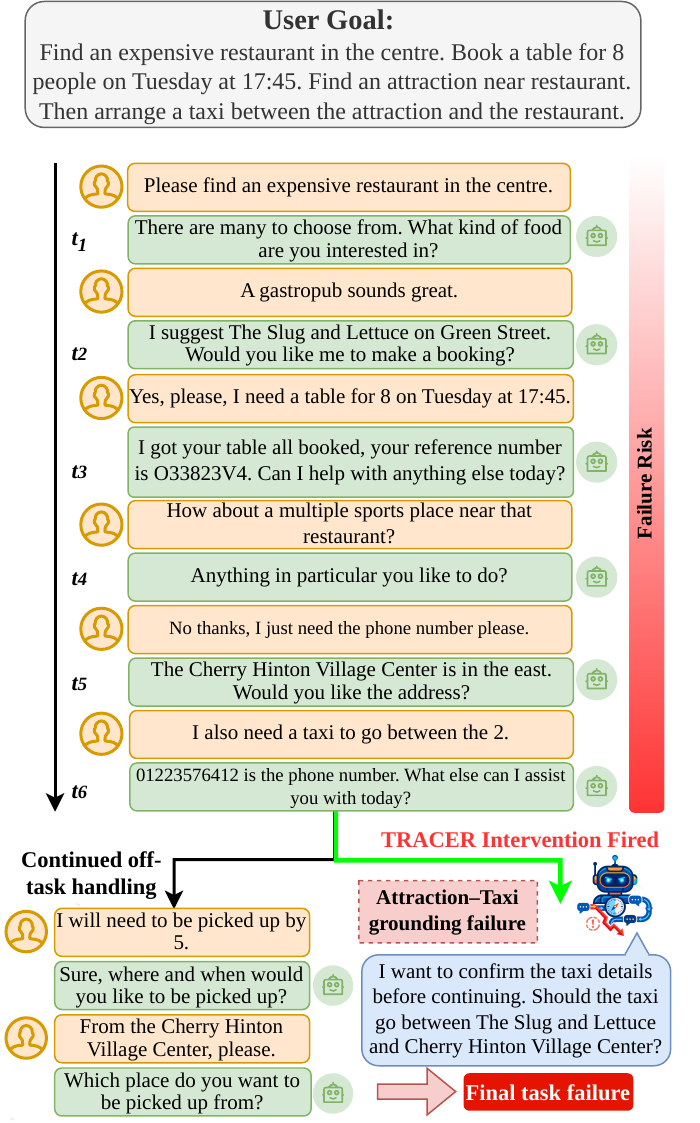}
    \caption{\textcolor{black}{Task overview. The example shows a conversation breaks down.}}
    \vspace{0em}
    %Illustrative failure trajectory in a multi-domain TOD dialogue. Although each turn is locally coherent, failure risk increases as the dialogue moves from restaurant booking to attraction search and taxi arrangement. The system books the restaurant successfully, but the later subtasks are not properly grounded: the attraction is not clearly connected to the booked restaurant, and the taxi endpoints remain ambiguous. TRACER therefore triggers a clarification-style intervention before the dialogue reaches final task failure
    \label{fig:tracer-motivation}
    \vspace{-1.8em}
\end{figure}

\section{Introduction}

Task-oriented dialogue (TOD) systems are designed to help users complete concrete goals, booking travel, finding restaurants, resolving customer-service queries, and they power an expanding range of commercial applications~\citep{young2013pomdp, wen2017network, qin2023end}. As these systems are deployed in ever-more-demanding settings, their ability to complete user goals reliably has become a practical engineering requirement that goes beyond benchmark scores~\citep{walker1997paradise, peng2021soloist}. Large-scale benchmarks such as MultiWOZ~\citep{budzianowski2018multiwoz}, the Schema-Guided Dialogue (SGD) dataset~\citep{rastogi2020towards}, and ABCD~\citep{chen2021abcd} have fueled rapid progress in dialogue state tracking (DST)~\citep{henderson2014second, wu2019trade, wu2020todbert}, end-to-end neural response generation~\citep{hosseini2020simple, lei2018sequicity}, and, more recently, LLM-based dialogue agents~\citep{hudecek2023llms}. Yet despite this progress, the dominant evaluation paradigm measures \emph{outcome} rather than \emph{trajectory}: whether the final state is correct, whether the final response is appropriate, or whether the overall task succeeded. For practical deployment this is often too late. %We further discover two complementary recovery methods, a template-based recovery module and a T5-based recovery generator, that operate under a shared trigger framework (see Appendix~\ref{sec:appendix-recovery-methods}).

As shown in Figure~\ref{fig:tracer-motivation}, many failures in TOD do not appear as a single abrupt mistake; they emerge \emph{progressively} through slot-value instability, unresolved contradictions, missing constraints, and loss of task focus. By the time the final outcome is measured, the dialogue may already have drifted too far for a lightweight intervention to help.

Several lines of research address related but distinct problems. PARADISE~\citep{walker1997paradise} established that task success rates underspecify dialogue quality, while breakdown detection~\citep{higashinaka2016dialogue} and TD-EVAL~\citep{acikgoz2025tdeval} target individual turn quality or post-hoc analysis of completed dialogues. In open-domain conversation, \citet{chang2019trouble} showed that sequential models over growing contexts can forecast derailment before breakdown, establishing partial-context prediction as principled and achievable. Clarification and repair~\citep{carletta1992planning, maier1997clarification, ngo2024exploration}, deep RL~\citep{li2016deep}, task pretraining~\citep{wu2020todbert}, and unified generative models~\citep{hosseini2020simple} have each improved dialogue quality or policy, but none defines intervention policies targeting trajectory-level failure.

Despite these advances, a critical gap remains: \emph{there is no principled framework for forecasting impending failure from partial task-oriented dialogue context and converting those forecasts into triggered recovery decisions}. Breakdown detection~\citep{higashinaka2016dialogue} predicts whether a target system utterance, given its preceding dialogue context, causes a dialogue breakdown. It does not forecast eventual task failure from partial dialogue prefixes or define an intervention policy over failure scores; post-hoc evaluators~\citep{walker1997paradise, acikgoz2025tdeval} analyze completed dialogues; derailment forecasting~\citep{chang2019trouble} relies on social signals rather than belief-state evolution (e.g., if a user is acting as troll); and LLM-based agents~\citep{hudecek2023llms} do not model failure trajectories or define thresholded intervention policies.

We present a dual-stream framework that addresses this gap by combining trajectory-level signals of belief-state evolution with serialized belief-state text to forecast impending dialogue failure from partial context. A development-tuned threshold converts failure scores into intervention decisions, enabling a principled study of early-warning timing and recovery quality.

This paper makes three contributions:
(1) \textcolor{black}{We formulate task-oriented dialogue failure prediction as a \emph{partial-dialogue forecasting} problem.} We introduce \textbf{TRACER}, a dual-stream model that predicts whether a task-oriented dialogue is likely to fail before the conversation ends. It combines trajectory features from dialogue-state changes with serialized belief-state text.
(2) We evaluate TRACER on multiple task-oriented dialogue benchmarks. The results show that useful failure signals appear from as little as 25\% of the dialogue, and that TRACER outperforms heuristic, classical, and single-stream baselines on early failure prediction.

  \begin{figure*}[h]
      \centering
      \includegraphics[width=\linewidth]{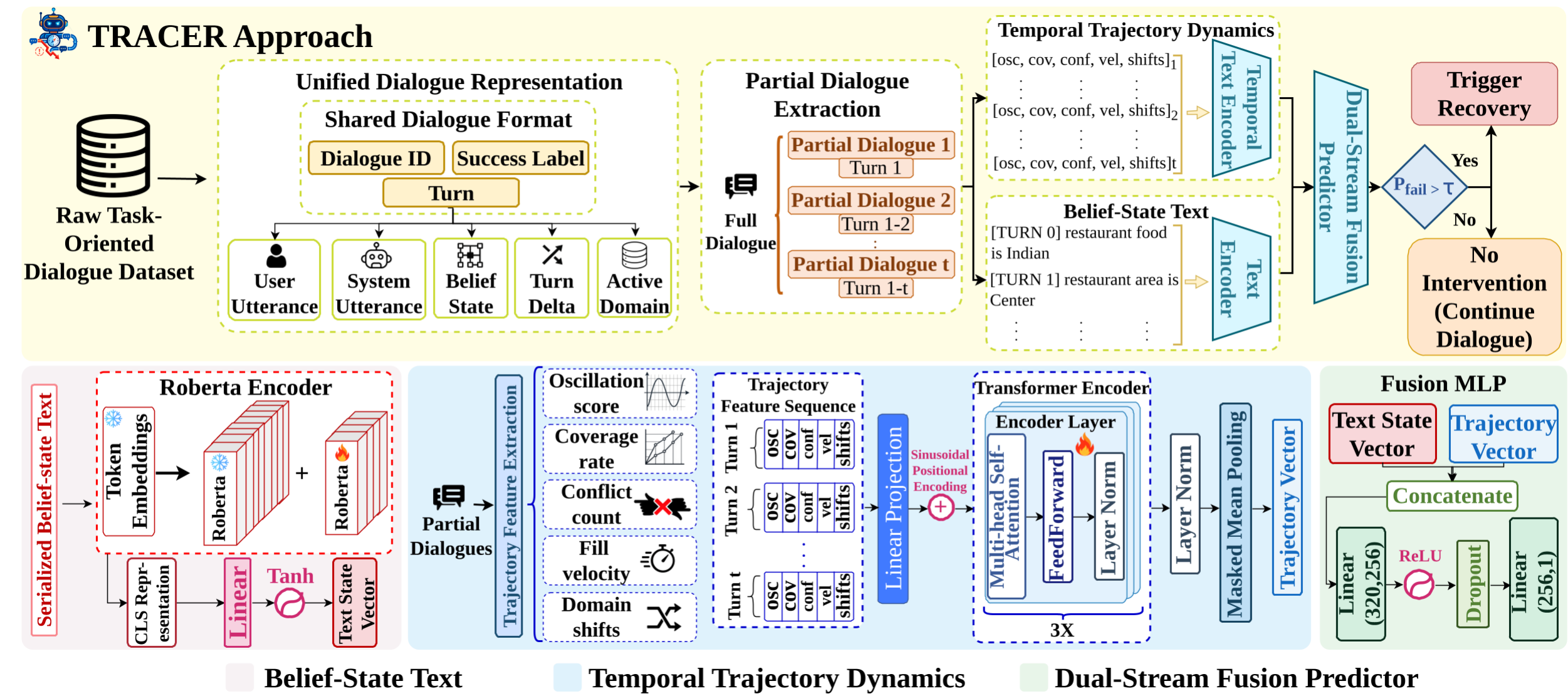}
  \caption{Overview of TRACER. Partial dialogues are encoded by a belief-state text stream and a temporal trajectory stream. The text stream uses RoBERTa to encode serialized belief states, while the trajectory stream models turn-level dialogue-state dynamics. Their representations are fused to predict failure risk and decide whether to trigger recovery.}
      \vspace{-1em}
      \label{fig:tracer-overview}
  \end{figure*}

\section{Related Work}

\paragraph{TOD systems, DST, and pretrained encoders.}
MultiWOZ~\citep{budzianowski2018multiwoz}, SGD~\citep{rastogi2020towards}, and ABCD~\citep{chen2021abcd} established multi-domain benchmarks for state tracking, response generation, and task completion. DST progress has been driven by cross-domain generalization~\citep{wu2019trade}, task-specific pretraining~\citep{wu2020todbert}, and sequence-to-sequence formulations~\citep{hosseini2020simple}, with end-to-end surveys~\citep{qin2023end} documenting the shift toward LLM-based agents~\citep{peng2021soloist, hudecek2023llms}. BERT~\citep{devlin2019bert} and RoBERTa~\citep{liu2019roberta} underpin strong dialogue encoders; TOD-BERT~\citep{wu2020todbert} extends this with dialogue-specific objectives. TRACER treats belief states as \emph{input}, using RoBERTa to encode their evolving content while a trajectory stream captures temporal dynamics not accessible from text alone.

\paragraph{Failure detection, forecasting, and repair.}
PARADISE~\citep{walker1997paradise} showed that dialogue quality cannot be reduced to a single end-of-dialogue label; the Dialogue Breakdown Detection Challenge~\citep{higashinaka2016dialogue} operationalized turn-level breakdown detection; and TD-EVAL~\citep{acikgoz2025tdeval} demonstrated that multi-granularity analysis exposes errors aggregate metrics miss. \citet{chang2019trouble} showed that sequential models over growing contexts can forecast conversational derailment before breakdown, establishing partial-context prediction as a principled framing. TRACER adapts this to TOD by grounding forecasting signals in belief-state evolution rather than social cues and couples prediction with a thresholded intervention policy. On the recovery side, clarification and repair are classical coherence mechanisms~\citep{carletta1992planning, maier1997clarification}\textcolor{black}{; \citet{mcroy1995repair} showed that abductive inference over mismatched speech acts can help identify/repair problems in task-oriented settings, an early computational treatment of failure}; recent work analyzing precursor cues~\citep{ngo2024exploration}; we build on this by asking whether an automated forecaster can supply a reliable trigger, comparing a template-based approach against a fine-tuned T5 generator~\citep{raffel2020exploring}.

\section{Methodology}

Figure~\ref{fig:tracer-motivation} shows the basic idea behind TRACER. In many task-oriented dialogues, failure does not come from a single bad turn. Instead, the dialogue gradually moves off track through state instability, contradictions, missing constraints, or domain confusion. Our goal is to detect this rising failure risk early, before the final breakdown becomes obvious, so that the system can decide whether to intervene. We therefore cast the problem as \emph{partial-dialogue forecasting}: given only the dialogue observed so far, the model predicts whether the full dialogue will eventually fail.

We formulate task-oriented dialogue failure prediction as a \emph{partial-dialogue-based forecasting} problem. Let a dialogue be $D = (u_1, s_1, \ldots, u_T, s_T)$, where $u_t$ and $s_t$ denote the user and system utterances at turn $t$, and let $y \in \{0,1\}$ denote the final dialogue outcome, with $y=1$ indicating eventual failure and $y=0$ indicating success. Instead of evaluating only on the full dialogue, we construct a supervised example for every visible partial dialogue $D_{\leq t}$, where the model observes turns $1{:}t$ and predicts whether the \emph{full} dialogue will eventually fail. This means that every dialogue contributes multiple training and test instances, one per partial dialogue instance, while all instances from the same dialogue share the same terminal label. Formally, let $\mathcal{D}$ be the full dialogue collection and let $T_i$ denote the length of dialogue $i$. The partial-dialogue dataset is:
$
\mathcal{D}_{\text{partial}} = \bigl\{(D_{\leq t},\; y_i) : i \in \mathcal{D},\; t \in \{1, \ldots, T_i\}\bigr\}$
where $y_i = 1$ if dialogue $i$ fails and $y_i = 0$ if it succeeds. A single dialogue of length $T_i$ contributes $T_i$ supervised examples, all sharing label $y_i$. The TRACER model learns a scoring function $f_\theta : D_{\leq t} \mapsto \hat{p}_{i,t} \in [0,1]$, and a threshold $\tau$ converts the continuous score into an intervention decision: trigger recovery when $\hat{p}_{i,t} > \tau$.

This setting is different from standard turn classification in two ways. First, the label is about the future. The goal is not to decide whether the current turn is problematic, but whether the dialogue is moving toward failure. Second, the prediction is meant to support intervention. To be useful in practice, the model’s score must be turned into a decision about when to act. Therefore, our method includes both a failure predictor and a threshold-based intervention policy. Figure~\ref{fig:tracer-overview}shows the overall pipeline. A partial dialogue is encoded using a trajectory stream and a belief-text stream. These representations are combined to predict a failure score, which triggers a recovery action when it exceeds a threshold tuned on the development set.

\subsection{Partial Dialogue Representation}

Each partial dialogue is represented using two complementary views of the dialogue state, corresponding to the two input streams shown in Figure~\ref{fig:tracer-overview}.

\paragraph{Trajectory features.}
As shown in the Temporal Trajectory Dynamics stream of Figure~\ref{fig:tracer-overview}, for every turn we compute a 5-dimensional trajectory vector that summarizes how the dialogue state is evolving up to that point: (1) oscillation score, the number of slots whose values change repeatedly over the dialogue history; (2) coverage rate, the fraction of required slots filled for the active domain; (3) conflict count, the cumulative number of detected contradictions between the user input and the current dialogue state; (4) fill velocity, the short-horizon rate at which new slot information is being filled; and (5) domain shifts, the number of distinct domains visited so far. A \textit{slot} is a something the system uses to remember one detail the user cares about, such as the restaurant's food type, price range, or location. For example, in ``find a cheap Italian restaurant downtown,'' the slots are \texttt{price = cheap}, \texttt{food = Italian}, and \texttt{area = downtown}. These are things mentioned in the conversation.
These features are computed incrementally for every partial dialogue, producing a sequence of turn-level vectors. This sequential view is important because many task failures emerge through unstable trajectories rather than through any single isolated error.

We define each feature formally. Let $b_t : \mathcal{S} \to \mathcal{V} \cup \{\emptyset\}$ be the cumulative belief state at turn $t$, where $\mathcal{S}$ is the set of domain-slot pairs and $\mathcal{V}$ is the value vocabulary. Let $d_t$ denote the active domain, $|b_t|$ the number of filled slots, and $\Delta_t = \{s \in \mathcal{S} : b_t(s) \neq b_{t-1}(s)\}$ the set of slots that changed at turn $t$.
\textit{Concretely,} $\mathcal{S}$ contains pairs such as \texttt{hotel-area} or \texttt{restaurant-food}, and $\mathcal{V}$ the corresponding values such as \texttt{north} or \texttt{italian}. The codomain $\mathcal{V} \cup \{\emptyset\}$ allows slots that have not yet been mentioned to map to $\emptyset$ (the unfilled state). For example, after three turns a belief state might read $b_3(\texttt{hotel-area}) = \texttt{north}$, $b_3(\texttt{hotel-pricerange}) = \emptyset$ (not yet specified), $b_3(\texttt{restaurant-food}) = \texttt{italian}$, and the change set $\Delta_3$ would contain only \texttt{restaurant-food} if that was the only slot updated at turn~3.

\paragraph{Oscillation Score.}
Let $V_s^t$ be the ordered sequence of distinct consecutive values assigned to slot $s$ through turn $t$. A slot oscillates when its value has been revised \textcolor{black}{at least twice, that is, when three or more distinct consecutive values have been assigned to it ($|V_s^t| > 2$)}:
\begin{equation*}
\text{OscScore}_t = \bigl|\{s \in \mathcal{S} : |V_s^t| > 2\}\bigr|
\label{eq:osc}
\end{equation*}
\textit{Example.} If a user first requests a \texttt{cheap} hotel, then revises to \texttt{expensive}, and then reverts to \texttt{cheap}, the slot \texttt{hotel-pricerange} accumulates three distinct consecutive values, so $|V_s^t| = 3 > 2$ and it contributes~1 to $\text{OscScore}_t$. A slot that is set once and never changed would have $|V_s^t| = 1$ and would not count.

\paragraph{Coverage Rate.}
Let $\mathcal{S}^*_d$ be the set of required informable slots for domain $d$. Coverage measures the fraction of required slots that are currently filled:
\begin{equation*}
\text{CoverageRate}_t = \frac{|\{s \in \mathcal{S}^*_{d_t} : b_t(s) \neq \emptyset\}|}{|\mathcal{S}^*_{d_t}|}
\label{eq:cov}
\end{equation*}
\textit{Example.} Suppose the hotel domain requires three informable slots: \texttt{area}, \texttt{pricerange}, and \texttt{type}. If at turn $t$ the user has specified \texttt{area = north} and \texttt{type = guesthouse} but has not yet mentioned a price range, then $\text{CoverageRate}_t = 2/3 \approx 0.67$. A dialogue that reaches its final turn without ever providing the price range would end with coverage below~1.0, signalling an incomplete task specification.

\paragraph{Conflict Count.}
A conflict at turn $\tau$ is detected when a slot value is overwritten and the user utterance is semantically distant from the prior value. Let $\phi(\cdot)$ denote sentence embeddings and $v_s^{\tau-1}$ the text \texttt{"The \{slot\} is \{old\_value\}"}:
\begin{equation*}
\resizebox{\columnwidth}{!}{$
c_\tau = \mathbf{1}\!\Bigl[\exists\, s \in \Delta_\tau \cap \operatorname{dom}(b_{\tau-1}) : \dfrac{\phi(u_\tau)^\top \phi(v_s^{\tau-1})}{\|\phi(u_\tau)\|\cdot\|\phi(v_s^{\tau-1})\|} < \delta\Bigr]
$}
\label{eq:conflict-indicator}
\end{equation*}
\begin{equation*}
\text{ConflictCount}_t = \sum_{\tau=1}^{t} c_\tau
\label{eq:conflict}
\end{equation*}
where $\delta = 0.3$ is the cosine-similarity threshold below which a change is flagged as conflicting.
\textit{Example.} Suppose the current belief state records \texttt{hotel-area = north} and at turn $\tau$ the user says ``actually, I would prefer the south side.'' The slot value changes to \texttt{south}, but the user utterance is semantically distant from the probe phrase ``The area is north'' (cosine similarity well below~0.3), so $c_\tau = 1$ and the conflict count increments. By contrast, if the user says ``yes, north works'' and the value stays unchanged, no slot appears in $\Delta_\tau$ and $c_\tau = 0$.

\paragraph{Fill Velocity.}
A rolling two-turn window measures the rate of new slot filling:
\begin{equation*}
\text{FillVelocity}_t = \begin{cases}
\dfrac{|b_t|}{t+1} & \text{if } t < 2 \\[6pt]
\dfrac{|b_t| - |b_{t-2}|}{2} & \text{otherwise}
\end{cases}
\label{eq:velocity}
\end{equation*}
\textit{Example.} If four slots are filled at turn $t{=}5$ and only two were filled at turn $t{=}3$, then $\text{FillVelocity}_5 = (4-2)/2 = 1.0$: the conversation is gathering new information at a steady pace. If at turn $t{=}7$ the count is still four (no new slots added since turn~5), then $\text{FillVelocity}_7 = 0$, signalling that the information exchange has stalled.

\paragraph{Domain Shifts.}
The count of distinct domains activated up to turn $t$:
$\text{DomainShifts}_t = \bigl|\{d_\tau : \tau \leq t,\; d_\tau \neq \emptyset\}\bigr|$

\textit{Example.} If a dialogue discusses a hotel in turns~1--3, switches to a restaurant query in turn~4, and adds a train booking in turn~6, then $\text{DomainShifts}_6 = 3$. A single-domain conversation that only ever touches \texttt{hotel} would have $\text{DomainShifts}_t = 1$ throughout its entire length.

\noindent The complete trajectory feature vector at turn~$t$ is defined as
$\mathbf{f}_t = (\text{OscScore}_t,\ \text{CoverageRate}_t,$
$\text{ConflictCount}_t,\ \text{FillVelocity}_t,\ \text{DomainShifts}_t)$
and lies in $\mathbb{R}^5$.

\paragraph{Belief-state text.}
In parallel, as shown in the Belief-State Text stream of Figure~\ref{fig:tracer-overview}, we serialize the belief state at each turn into text and concatenate the resulting descriptions across the visible partial dialogue. Concretely, each turn contributes a segment of the form \texttt{[TURN$t$] slot is value, ...}, and the partial dialogue text is the concatenation of all such segments up to turn $t$. This gives the model direct access to the semantic content of the evolving belief state, while keeping the input format agnostic to the exact ontology size.

\subsection{TRACER Architecture}

TRACER is a dual-stream predictor that combines a trajectory encoder with a belief-text encoder.

\paragraph{Stream A: trajectory encoder.}
As shown in Figure~\ref{fig:tracer-overview}, the trajectory stream projects each 5-dimensional turn vector into a latent representation, adds sinusoidal positional encodings to preserve turn order, and passes the sequence through a temporal Transformer encoder~\citep{vaswani2017attention}. Masked mean pooling over visible turns yields a fixed-dimensional trajectory representation $h_A$. In our experiments, this stream uses a 3-layer Transformer encoder with model dimension 64 and 4 attention heads. Formally, each feature vector $\mathbf{f}_t \in \mathbb{R}^5$ is projected and position-encoded:
\begin{equation*}
\resizebox{\linewidth}{!}{$
\mathbf{e}_t^A = \mathbf{W}_A \mathbf{f}_t + \mathbf{p}_t, \quad \mathbf{W}_A \in \mathbb{R}^{d_A \times 5},\; d_A = 64
$}
\label{eq:stream-a-proj}
\end{equation*}
where $\mathbf{p}_t$ is the sinusoidal positional encoding at position $t$. The projected sequence $(\mathbf{e}_1^A, \ldots, \mathbf{e}_t^A)$ is processed by a 3-layer multi-head self-attention Transformer with $H=4$ heads, and the outputs at valid positions are mean-pooled:
\begin{equation*}
\resizebox{\linewidth}{!}{$
h_A = \frac{1}{t} \sum_{\tau=1}^{t} \bigl[\mathrm{TransformerEncoder}_A(\mathbf{e}_{1:t}^A)\bigr]_\tau \in \mathbb{R}^{64}
$}
\label{eq:stream-a-pool}
\end{equation*}

\paragraph{Stream B: belief-text encoder.}
As shown in the Belief-State Text stream of Figure~\ref{fig:tracer-overview}, the text stream encodes the serialized belief-state text with a pretrained RoBERTa encoder~\citep{liu2019roberta}. We use the contextualized first-token representation and project it into a 256-dimensional belief-state embedding $h_B$. To stabilize training, the lower encoder layers are partially frozen, while the upper layers and projection head remain trainable. Formally, let $x_t$ be the serialized belief-state text for the partial dialogue up to turn $t$:
\begin{equation*}
\resizebox{\linewidth}{!}{$
h_B = \tanh\!\bigl(\mathbf{W}_B \cdot \mathrm{RoBERTa}(x_t)_{[\mathrm{CLS}]}\bigr) \in \mathbb{R}^{256}, \quad \mathbf{W}_B \in \mathbb{R}^{256 \times 768}
$}
\label{eq:stream-b}
\end{equation*}
where $\mathrm{RoBERTa}(\cdot)_{[\mathrm{CLS}]}$ denotes the contextualized representation of the first token.

\paragraph{Fusion and prediction.}
As shown in the Fusion MLP of Figure~\ref{fig:tracer-overview}, the final TRACER representation is the concatenation $[h_A;\, h_B] \in \mathbb{R}^{320}$, which is passed through a small multilayer perceptron with ReLU activation and dropout to produce a scalar logit. After a sigmoid transformation, this yields $\hat{p}(y=1 \mid D_{\leq t})$, the probability that the dialogue will eventually fail:
\begin{equation*}
\resizebox{\linewidth}{!}{$
\hat{p}(y=1 \mid D_{\leq t}) = \sigma\!\Bigl(\mathbf{W}_3\,\mathrm{ReLU}\bigl(\mathbf{W}_2\,[h_A;\,h_B] + \mathbf{b}_2\bigr) + b_3\Bigr)
$}
\label{eq:fusion}
\end{equation*}
where $\mathbf{W}_2 \in \mathbb{R}^{256 \times 320}$, $\mathbf{W}_3 \in \mathbb{R}^{1 \times 256}$, and $\sigma(\cdot)$ is the sigmoid function.

\subsection{Training Objective}

TRACER is trained with a binary classification objective over partial dialogues, combined with an early-prediction reward that upweights correct failure predictions made at earlier turns. Intuitively, a correct forecast early in a dialogue is more valuable than one made near the end, when intervention opportunities are limited. This objective reflects the practical goal of the system: not merely to classify dialogues accurately, but to surface risk early enough to be actionable.

Formally, let $\mathcal{P}_{\text{train}} = \{(i, t) : i \in \mathcal{D}_{\text{train}},\; t \in \{1, \ldots, T_i\}\}$ be the set of all partial-dialogue training instances. The combined loss is:
\begin{equation*}
\resizebox{\linewidth}{!}{$
\mathcal{L}_{\text{BCE}} = -\frac{1}{|\mathcal{P}_{\text{train}}|}\sum_{(i,t) \in \mathcal{P}_{\text{train}}} \Bigl[y_i \log \hat{p}_{i,t} + (1-y_i)\log(1-\hat{p}_{i,t})\Bigr]
$}
\label{eq:bce}
\end{equation*}

\noindent The early-prediction reward term additionally encourages the model to fire correct failure warnings as early as possible, weighted by how much dialogue remains:
\begin{equation*}
\resizebox{\linewidth}{!}{$
\mathcal{L}_{\text{early}} = -\frac{\lambda}{|\mathcal{P}_{\text{train}}|}\sum_{(i,t) \in \mathcal{P}_{\text{train}}} w(t, T_i)\cdot\mathbf{1}\!\bigl[\hat{p}_{i,t} > \tfrac{1}{2},\; y_i = 1\bigr]
$}
\label{eq:early}
\end{equation*}

\noindent The two terms combine into the total training loss:
$
\mathcal{L} = \mathcal{L}_{\text{BCE}} + \mathcal{L}_{\text{early}}
\label{eq:total-loss}
$
where $w(t, T_i) = (T_i - t)/T_i \in [0,1]$ is a temporal weight assigning higher reward to correct failure predictions made at earlier turns, and $\lambda = 0.3$. The reward term is active only for true-positive partial-dialogue predictions (failed dialogues where the model correctly assigns $\hat{p}_{i,t} > 0.5$). \textcolor{black}{Crucially, $w(t, T_i)$ uses the known full-dialogue length $T_i$, which is available during \emph{training} because the supervision is derived from completed dialogues. At inference time, $\mathcal{L}_{\text{early}}$ is not computed; the model simply outputs $\hat{p}_{i,t}$ from the observed partial dialogue without requiring knowledge of the future dialogue length.} To address class imbalance from partial-dialogue expansion, the binary cross-entropy applies positive-class upweighting with weight $\alpha_+ = |\mathcal{P}^-_{\text{train}}|/|\mathcal{P}^+_{\text{train}}|$. Full experimental details are provided in Appendix~\ref{sec:appendix-experimental-setup}.

\begin{table*}[t]
    \centering
    \small
    \setlength{\tabcolsep}{5.5pt}
    \renewcommand{\arraystretch}{1.15}
    \begin{tabular}{llccccc}
        \toprule
        \textbf{Category} & \textbf{Model} & \textbf{Belief State} 
            & \textbf{AUC-ROC} $\uparrow$ & \textbf{F1} $\uparrow$ 
            & \textbf{EDS} $\uparrow$ & \textbf{Det. Turn} $\downarrow$ \\
        \midrule
        \multirow{4}{*}{Heuristic Trigger}
            & No intervention             & oracle & 0.500 & 0.000 & 0.000 & $\infty$ \\
            & Fixed-schedule (every 3)    & oracle & 0.499 & 0.236 & 0.690 & 2.00     \\
            & Slot-confidence (Osc$>$0)   & oracle & 0.495 & 0.025 & 0.009 & 7.50     \\
            & Feature-threshold ensemble  & oracle & 0.535 & 0.286 & 0.864 & 0.25     \\
        \midrule
        \multirow{2}{*}{\shortstack[l]{Classical\\Feature Model}}
            & Partial dialogue logreg (turn-level)  & oracle & 0.574 & 0.344 & 0.849 & 0.35 \\
            & Partial dialogue logreg (engineered)  & oracle & 0.600 & 0.351 & 0.728 & 1.03 \\
        \midrule
        \multirow{3}{*}{\shortstack[l]{Zero-Shot\\Prompting}}
            & Llama-3.1-8B  & oracle & 0.519 & 0.268 & 0.474 & 2.59 \\
            & Mistral-7B    & oracle & 0.525 & 0.278 & 0.558 & 2.08 \\
            & Qwen2.5-7B    & oracle & 0.531 & 0.321 & 0.772 & 1.32 \\
        \midrule
        \multirow{3}{*}{\shortstack[l]{Zero-Shot CoT\\Prompting}}
            & Llama-3.1-8B  & oracle & 0.516 & 0.290 & 0.566 & 2.49 \\
            & Mistral-7B    & oracle & 0.526 & 0.275 & 0.568 & 1.95 \\
            & Qwen2.5-7B    & oracle & 0.519 & 0.318 & 0.756 & 1.34 \\
        \midrule
        \multirow{3}{*}{\shortstack[l]{Few-Shot\\Prompting}}
            & Llama-3.1-8B  & oracle & 0.492 & 0.337 & 0.993 & 0.06 \\
            & Mistral-7B    & oracle & 0.524 & 0.312 & 0.648 & 1.89 \\
            & Qwen2.5-7B    & oracle & 0.518 & 0.299 & 0.668 & 1.67 \\
        \midrule
        \multirow{2}{*}{\textbf{Our Approach}}
            & \textbf{TRACER} & generated & \underline{0.617} & \underline{0.370} & 0.751 & 1.22 \\
            & \textbf{TRACER} & oracle    & \textbf{0.741}$^{\textcolor{black}{\dagger}}$    & \textbf{0.463}$^{\textcolor{black}{\dagger}}$    & 0.727 & 1.36 \\
        \bottomrule
    \end{tabular}

      \caption{Main failure-forecasting results on MultiWOZ. TRACER performs best overall, though some heuristic baselines obtain higher EDS through aggressive triggering. \textcolor{black}{$^\dagger$Significantly better than both the best classical baseline (logreg engineered) and best LLM baseline (Qwen2.5-7B zero-shot): $p < 0.0001$ by a paired bootstrap test.}}
    
    \vspace{-1.5em}
    \label{tab:main-forecasting}
\end{table*}

\subsection{Triggered Recovery Policy}

The predictor becomes a control layer by thresholding the partial-dialogue-level failure score. For a partial dialogue at turn $t$, if $p(y=1 \mid D_{\leq t}) > \tau$, the system triggers a recovery action. We select $\tau$ on the development set under a false-positive-rate constraint and then freeze that operating point for test-time evaluation. This distinction matters because unconstrained low thresholds can maximize early-detection metrics trivially while producing an impractically large number of unnecessary interventions. We study two recovery methods under this shared trigger policy; their design and implementation are described in Appendix~\ref{sec:appendix-recovery-methods}.

%\subsection{Scope of the Method}

%TRACER is designed to evaluate whether impending failure can be detected from partial dialogue context and whether those forecasts are useful as intervention triggers. It is therefore not a full online dialogue manager, and our recovery analysis should be interpreted accordingly. The recovery modules are evaluated offline under a shared trigger policy, which lets us compare intervention quality and cost without claiming direct causal improvements in downstream task success. This framing is deliberate: the method contribution of TRACER is the forecasting-and-triggering framework shown in Figure~\ref{fig:tracer-overview}, while recovery quality and qualitative failure taxonomy are supporting analyses built on top of that control signal. 

\section{Results}

\noindent \textbf{Evaluation Settings.} We evaluate TRACER in both an \emph{oracle} setting, where gold belief states are provided, and a \emph{generated} setting, where belief states are produced automatically to better reflect deployment conditions. This comparison lets us separate the forecasting capacity of the model under clean structured inputs from its robustness when upstream state representations are noisy. Details of the belief-state generation prompting procedure are provided in Appendix~\ref{sec:prompt-belief}.
We report four metrics. Two standard metrics: \textbf{AUC-ROC}, \textbf{False Positive Rate (FPR)}, and \textbf{F1} (with a threshold set via the  development-set). \textbf{Early Detection Score (EDS)} measures how early the model first correctly triggers on a failing dialogue: $\text{EDS}_i = (T_i - t_i^*)/T_i$, where $t_i^*$ is the first trigger turn; EDS $= 1$ for a turn-1 detection and EDS $= 0$ if the model never triggers. \textbf{Det.\ Turn} ($\bar{t}^*$) is the mean turn at which the first correct trigger fires, averaged over detected failing dialogues; lower is earlier.  Formal definitions are in Appendix~\ref{sec:appendix-metrics}.

\textcolor{black}{\noindent \textbf{Baselines.} We compare against three groups. \emph{Heuristic triggers} include a no-intervention baseline, a fixed-schedule trigger (fires every 3 turns), a slot-confidence trigger (fires on any oscillation), and a feature-threshold ensemble (fires when any hand-designed threshold over the five trajectory features is exceeded). \emph{Classical feature models} are two logistic-regression classifiers trained on trajectory features, one using the turn-level feature vector and one using an engineered partial-dialogue summary. \emph{Prompting baselines} evaluate Llama-3.1-8B, Mistral-7B-Instruct, and Qwen2.5-7B under zero-shot, zero-shot chain-of-thought, and few-shot paradigms. Full baseline details are in Appendix~\ref{sec:appendix-experimental-setup}.}

\noindent \textbf{Main Forecasting Performance.}
Table~\ref{tab:main-forecasting} shows the main partial-dialogue-based forecasting results on the MultiWOZ test split. TRACER achieves the strongest discrimination performance overall, with an AUC-ROC of 0.741 and an F1 score of 0.463. This is a clear improvement over the strongest learned single-stream baseline, the text-only model, which achieves 0.692 AUC and 0.432 F1, and a larger gain over classical engineered partial-dialogue baselines and hand-designed trigger heuristics. These results support the paper's central modeling claim: combining trajectory dynamics with belief-state text is more effective for forecasting eventual failure than relying on either source alone.

At the same time, Table~\ref{tab:main-forecasting} also shows why forecasting quality cannot be reduced to a single early-trigger metric. Some heuristic baselines achieve high EDS by firing extremely aggressively. For example, the feature-threshold ensemble attains an EDS of 0.864 despite much weaker ranking and classification quality (0.535 AUC, 0.286 F1), and the fixed-schedule baseline reaches 0.690 EDS while remaining close to chance in AUC. TRACER's advantage lies in producing a more reliable failure score, not merely in triggering as early as possible, a distinction examined in Section~\ref{subsec:threshold-tradeoff}. {\color{black}
Several methods in Table~\ref{tab:main-forecasting} report mean detection turns below 3, which may seem surprising for MultiWOZ dialogues that can exceed 10 turns. The key is that Det.\ Turn is the mean \emph{first} trigger turn, averaged only over detected failing dialogues. Aggressive baselines (e.g., the feature-threshold ensemble with Det.\ Turn $= 0.25$) fire almost immediately on nearly every dialogue, which yields a near-zero mean detection turn but also produces extremely high false-positive rates. TRACER's Det.\ Turn of 1.36 reflects a threshold calibrated under the FPR $\leq 0.15$ constraint, so it triggers early but selectively.

\noindent \textbf{Relationship between Det.\ Turn in Table~\ref{tab:main-forecasting} and Fixed-Context Forecasting.}
These are two distinct analyses. Table~\ref{tab:main-forecasting} reports the \emph{threshold-triggered} behavior: the model sees turns one by one and fires once the failure score crosses $\tau$, so Det.\ Turn is the earliest crossing point. Fixed-Context Forecasting (Section below) instead \emph{locks} the visible context to a fixed fraction of the dialogue (25\%, 50\%, 75\%, or 100\%) and evaluates classification and ranking quality at that snapshot; no threshold trigger is involved. The two analyses are complementary: one measures \emph{when} the system would intervene, while the other measures \emph{how discriminative} partial information is at different stages.
}

\textcolor{black}{\noindent \textbf{Oracle vs.\ generated belief states.} TRACER with oracle belief states achieves higher AUC-ROC (0.741 vs.\ 0.617) and F1 (0.463 vs.\ 0.370) than with generated states, as expected from cleaner supervision. However, the generated setting sometimes shows marginally higher EDS (0.751 vs.\ 0.727). This counter-intuitive pattern arises because the development-set threshold is re-calibrated for each belief-state condition, and the noisier generated belief states shift the model's score distribution in ways that can move the frozen threshold to a slightly more aggressive trigger point. It does not indicate that generated states provide better information; AUC-ROC, which is threshold-independent, is the appropriate summary of discrimination quality.}

{\color{black}

}

\vspace{1mm}
\noindent \textbf{Ablation Results.}
\label{sec:results-ablation}
Table~\ref{tab:ablations} isolates the contribution of the two input streams under both oracle and generated belief-state inputs. With oracle belief states, the text-only model is stronger than the features-only model, indicating that the serialized belief-state text already contains substantial predictive signal about eventual failure. With generated belief states, both single-stream ablations degrade, but the full fused model remains strongest, showing that trajectory dynamics and belief-state text remain complementary even when the belief state is noisier.

\begin{table}[t]
    \centering
    \small
    \setlength{\tabcolsep}{0pt}
    \begin{tabularx}{\columnwidth}{>{\centering\arraybackslash}X>{\centering\arraybackslash}Xcc}
        \toprule
        \textbf{Model} & \textbf{Belief State Type} & \textbf{AUC-ROC} $\uparrow$ & \textbf{F1} $\uparrow$ \\
        \midrule
        \multirow{2}{*}{\makecell{Stream A only \\ (features)}}
            & generated & 0.554             & 0.337 \\
            & oracle    & 0.623             & 0.363 \\
        \midrule
        \multirow{2}{*}{\makecell{Stream B only \\ (text)}}
            & generated & 0.583             & 0.332 \\
            & oracle    & \underline{0.692} & \underline{0.432} \\
        \midrule
        \multirow{2}{*}{\textbf{TRACER}}
            & generated & 0.617             & 0.370 \\
            & oracle    & \textbf{0.741}    & \textbf{0.463} \\
        \bottomrule
    \end{tabularx}
    \caption{Ablation on TRACER streams under oracle and generated belief-state inputs.
    The fused model outperforms both single-stream ablations in both settings.}
    \vspace{-1.5em}
    \label{tab:ablations}
\end{table}

\begin{figure}[t]
    \centering
    \includegraphics[width=\linewidth]{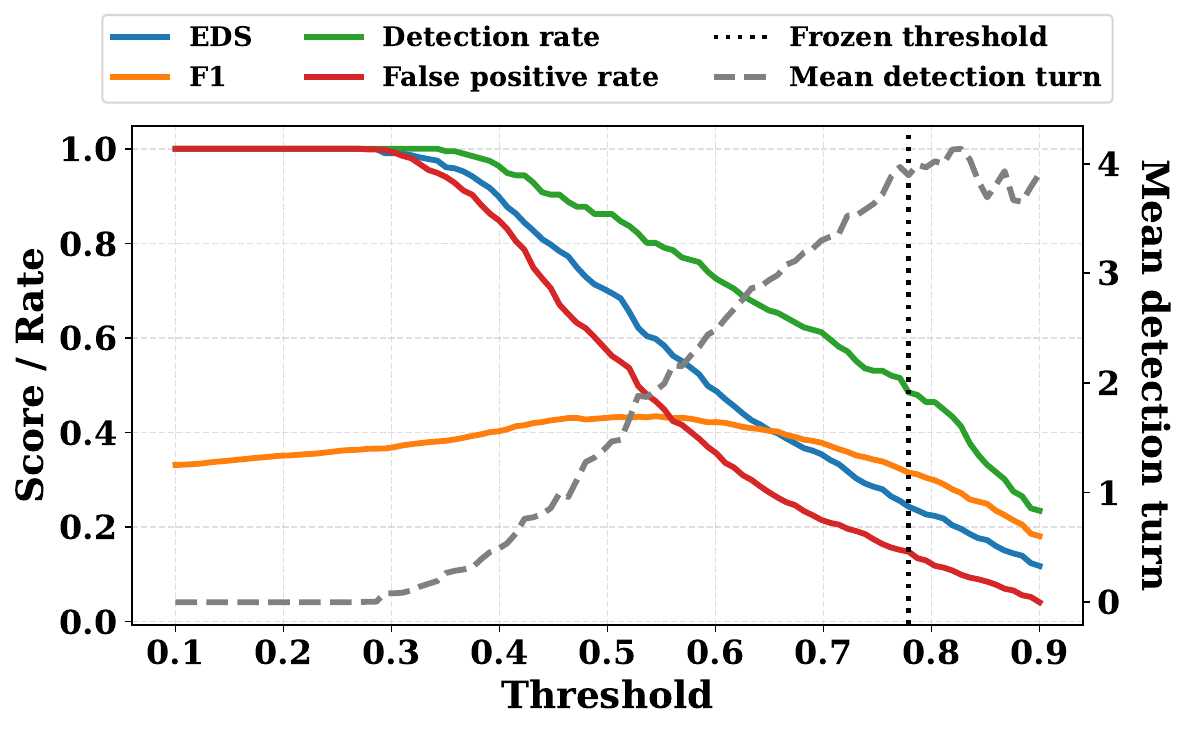}
    \caption{Threshold tradeoff on the development set. The figure should show how lower thresholds improve early detection but sharply increase false positives.}
    \vspace{-1.5em}
    \label{fig:threshold-tradeoff}
\end{figure}

\vspace{1mm}
\noindent \textbf{Threshold Tradeoff Analysis.}
\label{subsec:threshold-tradeoff}
The raw forecasting scores become operational only after they are converted into intervention decisions through a threshold. We select this threshold on the \textbf{development set} (distinct from the test set used for final results) under an explicit false-positive-rate constraint, so that the operating point reflects a principled choice rather than post-hoc tuning on test data. Figure~\ref{fig:threshold-tradeoff} plots the threshold sweep on the development set. The main pattern is clear: lower thresholds detect more failures earlier, but they also create many false positives. At the extreme, the unconstrained best-EDS point occurs at $\tau = 0.1$, where the detection rate reaches 1.0, but the false positive rate also reaches 1.0, making the operating point unusable in practice.

We therefore freeze a threshold selected under an explicit false-positive-rate constraint. The final operating point, $\tau = 0.7788$, yields a test false positive rate of 0.141, a detection rate of 0.540, mean detection turn 3.70, and F1 of 0.374. This is substantially more conservative than the threshold that maximizes development-set F1 ($\tau \approx 0.544$), which increases recall and F1 but also drives the false positive rate to roughly 0.47 on the development set. The threshold study therefore supports a key claim of the paper: EDS alone is not sufficient for choosing an intervention policy, and practical dialogue forecasting requires an explicit tradeoff between early warning and unnecessary interventions. \textcolor{black}{More precisely, the intervention threshold encodes a cost-benefit decision: a lower $\tau$ warns earlier but interrupts more successful dialogues (higher FPR), while a higher $\tau$ is more precise but misses more failures or catches them late. In settings where an unnecessary intervention is low-cost, such as a soft confirmation prompt, a more aggressive threshold may be preferable. In settings where interrupting a user mid-task carries higher friction, a conservative threshold like the one we freeze ($\tau = 0.7788$) is more appropriate. We report the full threshold sweep in Appendix~\ref{sec:appendix-forecasting}.}

\begin{figure}[t]
    \centering
    \includegraphics[width=\linewidth]{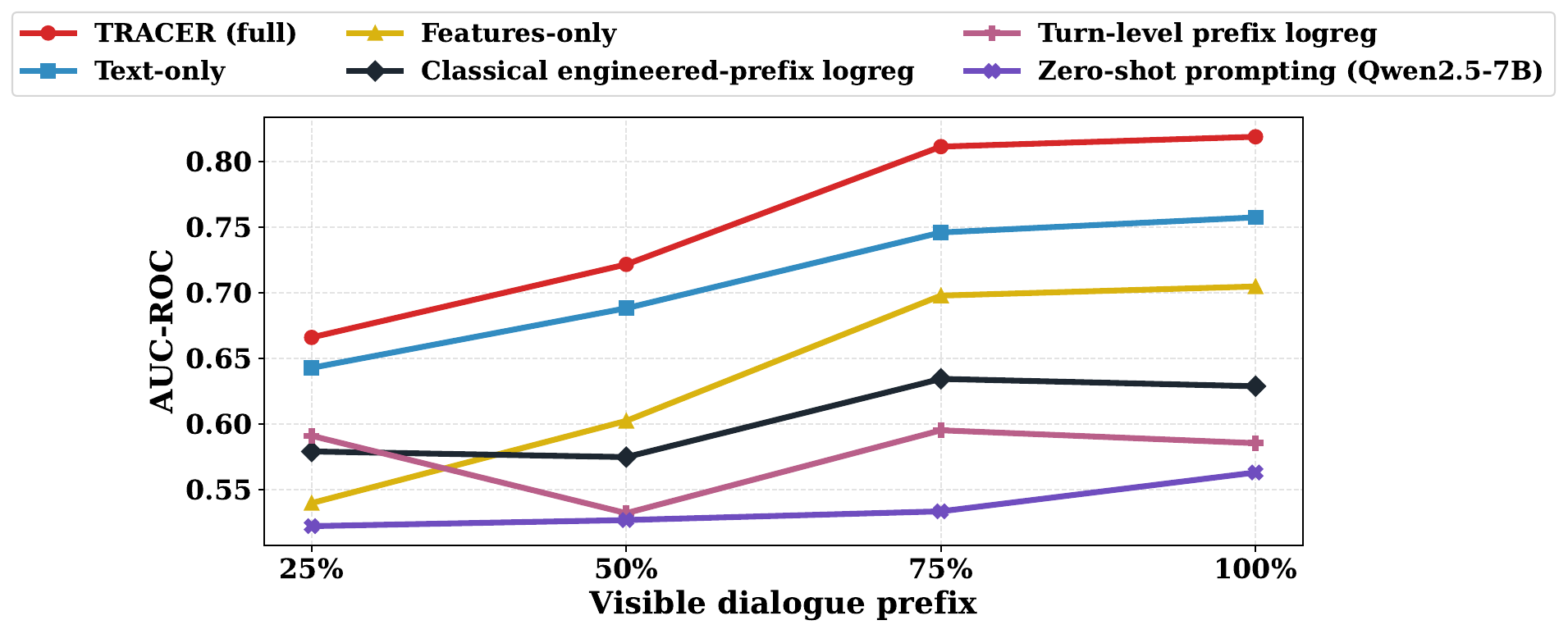}
    \caption{Fixed-context forecasting: TRACER recovers strong predictive signal from partial context, with substantial gains at 50\% and 75\% of the dialogue.}
    \vspace{-1em}
    \label{fig:partial-analysis}
\end{figure}

\vspace{1mm}
\noindent \textbf{Fixed-Context Forecasting.}
Figure~\ref{fig:partial-analysis} evaluates TRACER under fixed visible-context conditions, where only the first 25\%, 50\%, 75\%, or 100\% of each dialogue is revealed. This analysis makes the early-forecasting claim more concrete than aggregate EDS alone. Even with only the first quarter of the dialogue visible, TRACER already achieves an AUC-ROC of .666, indicating that a meaningful failure signal is available well before task completion. Performance improves steadily as more context is revealed: AUC rises to .722 at 50\%, .811 at 75\%, and 0.819 at full context, while F1 increases from .209 to .317, .461, and .524, respectively.

The most notable pattern is that by 75\% of the dialogue, TRACER is already close to its full-context discrimination quality. This suggests that impending failure is often visible before the final turns, which is exactly the regime where proactive intervention is plausible. Detection rate also increases monotonically from 0.141 at 25\% to 0.540 at \textcolor{black}{100\% (i.e., the complete dialogue), where the frozen threshold is applied to the full test set}, while EDS rises from 0.103 to 0.299. \textcolor{black}{F1 at 25\% is modest (0.209), which reflects the expected difficulty of predicting eventual task outcome from very limited context: at this stage most slot values have not yet been mentioned and the trajectory features carry little signal. This is consistent with the monotonic improvement in AUC-ROC across context lengths, which shows that the model's ranking quality improves steadily rather than steeply only near the end.} Taken together, these results provide direct evidence that the forecasting problem is not purely retrospective: partial dialogues contain enough signal to support useful early-warning behavior.

\vspace{1mm}
\noindent \textbf{Cross-Domain Transfer.}
Table~\ref{tab:cross-domain} in Appendix~\ref{sec:appendix-forecasting} reports zero-shot transfer from a MultiWOZ-trained TRACER model to SGD and ABCD. The results are mixed. On ABCD, TRACER transfers reasonably well, reaching 0.771 AUC-ROC, 0.772 F1, and 0.819 EDS, which suggests that the forecasting signal learned from trajectory dynamics and belief-state text can generalize to at least some task-oriented domains beyond the source benchmark. In contrast, zero-shot transfer to SGD is weak: AUC falls to 0.494 and F1 to 0.069, even though EDS remains moderately high. These results indicate that cross-domain generalization is non-uniform: TRACER transfers well to ABCD but fails on SGD, positioning it as a strong in-domain forecaster with partial transfer ability rather than a universally portable predictor. %\textcolor{black}{We attribute the ABCD success to structural similarity: ABCD customer-service dialogues follow booking-like task flows where slot coverage, oscillation, and fill velocity carry analogous meanings to MultiWOZ hotel/restaurant tasks. SGD transfer fails for three compounding reasons: (1) SGD's success labels are derived from system action annotations rather than user goal satisfaction, creating a different failure definition; (2) SGD covers 16+ services with highly heterogeneous schema, so the five trajectory features carry different distributional properties; and (3) the absence of fine-tuning or threshold recalibration on SGD means the frozen operating point is mis-calibrated for the new score distribution. Fine-tuning on even a small SGD sample, or schema-aware feature normalization, are promising directions for improving transfer.}

\vspace{1mm}
\noindent \textbf{Qualitative Error Analysis.}
Error analysis reveals two characteristic patterns. False positives often involve dialogues that are ultimately successful but pass through genuine instability, domain confusion, repeated corrections, or cross-domain shifts, that TRACER correctly flags as risky before the dialogue self-corrects. Not every false positive is semantically uninformative: the system detects real risk in dialogues that happen to recover. False negatives show the opposite: some failed dialogues remain locally coherent at the feature level until the final turns, making failure genuinely difficult to anticipate from partial context.

\vspace{1mm}
\noindent \textbf{Human-Evaluated Failure Taxonomy Analysis.}
To interpret TRACER failures, we define a 4-category taxonomy: \textit{Information Incompleteness} (task constraints never stably assembled), \textit{Contradiction / Misalignment} (system-user misalignment on facts or constraints), \textit{State Instability} (belief states oscillating without convergence), and \textit{Task Drift / Domain Confusion} (interactions shifting away from the active task or mixing incompatible domains). These labels are \emph{not} ground-truth annotations, but serve as a human-evaluated qualitative framework over sampled failed dialogues. Full definitions are in Appendix~\ref{sec:appendix-taxonomy}.

To validate this taxonomy, we annotate a balanced sample of 60 failed dialogues (15 per category); Table~\ref{tab:taxonomy-summary} reports heuristic and human-evaluated on-diagonal counts. Alignment is strongest for \textit{Contradiction / Misalignment} and \textit{Task Drift / Domain Confusion} (15/15 each), while \textit{Information Incompleteness} and \textit{State Instability} show minor leakage (2 off-diagonal each), as missing information and repeated slot revisions frequently co-occur. %\textcolor{black}{The 15/15 perfect alignment on contradiction and drift categories reflects the fact that these failure types produce distinctive surface signals (explicit user corrections; domain changes in the active-domain annotation) that the heuristic labeler and human annotators both respond to in the same way. The 13/15 partial alignment on incompleteness and instability is expected and not problematic: dialogues in those two categories often co-exhibit missing slots \emph{and} repeated revisions, so annotator disagreement arises from genuine category overlap rather than annotation error.} These labels serve as qualitative interpretive evidence that TRACER responds to recognizable failure patterns.

\begin{table}[t]
    \centering
    \small
    \begin{tabular}{lcc}
        \hline
        \textbf{Label} & \textbf{Heur.} & \textbf{On-Diag.} \\
        \hline
        Information Incompleteness    & 15 & 13 \\
        Contradiction / Misalignment  & 15 & 15 \\
        State Instability             & 15 & 13 \\
        Task Drift / Domain Confusion & 15 & 15 \\
        \hline
    \end{tabular}
    \caption{Human-evaluated qualitative taxonomy on a balanced 60-dialogue sample. \textbf{Heur.}: heuristic count; \textbf{On-Diag.}: human-evaluated on-diagonal count.}
    \vspace{-1em}
    \label{tab:taxonomy-summary}
\end{table}

\section{Conclusion}

We presented TRACER, a framework for early failure detection in task-oriented dialogue. Instead of waiting until the end of the conversation, TRACER predicts from partial dialogue context whether the interaction is likely to fail. It combines trajectory dynamics with belief-state text, which lets it capture both how the dialogue state changes over time and what the system currently believes the user wants. In our experiments, this approach detects useful failure signals well before dialogue completion and outperforms heuristic, classical, and single-stream baselines. These results suggest that early failure detection can serve as a practical control layer for task-oriented dialogue systems, complementing standard post-hoc evaluation methods. In future work, we plan to test TRACER in interactive settings with live recovery, improve transfer across datasets, and further validate the failure categories with human annotation.

\section*{Acknowledgments}
This material is based upon work supported by the National Science Foundation (NSF) under Grant~No. 2145357.

\newpage
\section*{Limitations}

TRACER shows that it is possible to predict dialogue failure from partial conversations, and that performance improves when the model uses both trajectory features and belief-state text. Still, the method has several limitations. First, the recovery analysis is offline, so the results do not show that TRACER improves task success in live conversations with real users. For example, a recovery action might look helpful in an offline evaluation but still confuse a real user or interrupt a conversation at the wrong time. This limitation is important because a dialogue system is only useful in practice if it helps real users complete their tasks.

Second, the failure taxonomy is only used to help interpret errors, not as gold-standard supervision, because public task-oriented dialogue datasets do not include these labels. For example, a dialogue may be labeled as showing task drift or state instability by our taxonomy, but that label is an interpretation rather than an official dataset annotation. This is still important because the taxonomy helps explain what kinds of failures the model is detecting, even if it is not a formal benchmark target.

Third, cross-domain transfer is inconsistent: the model transfers well to ABCD but poorly to SGD. For example, a model trained on MultiWOZ may still detect failure patterns in ABCD, but fail to generalize to SGD because the dialogue structure, domains, or failure types differ. This matters because a method is much more useful if it works across datasets rather than only on the domain it was trained on.

\textcolor{black}{Fourth, some trajectory features are intentionally coarse. The DomainShifts feature counts the number of \emph{distinct} domains activated so far, rather than modeling actual sequential domain transitions. A dialogue that visits hotel, restaurant, and hotel again registers the same DomainShifts value as one that visits three entirely new domains in sequence. This simplification makes the feature easy to compute incrementally and correlates with multi-domain failure risk, but it does not capture the directionality or recency of domain switches. A proper transition-count or recency-weighted domain-shift feature could capture more subtle task-drift patterns and is a direction for future work.}

Finally, EDS by itself is not enough for evaluation, since a system can get a high EDS simply by triggering too often and creating too many false positives. For example, a model could warn very early in almost every conversation, which would improve EDS but would also lead to many unnecessary interventions in successful dialogues. This is important because a good intervention system must balance early warning with not interrupting users when nothing is wrong.

Future work should test the model in interactive settings, improve transfer across datasets, and validate the failure categories more carefully with human annotation.

\bibliography{custom}

\appendix

\section{Experimental Setup}
\label{sec:appendix-experimental-setup}

\subsection{Datasets}

We train and evaluate TRACER primarily on MultiWOZ 2.4~\citep{budzianowski2018multiwoz}, which provides multi-domain dialogues with explicit belief states and rich opportunities for gradual failure through missing information, slot inconsistency, or domain drift. Partial-dialogue supervision is derived from terminal dialogue outcomes, with trajectory features computed turn by turn. To test generalization, we evaluate the MultiWOZ-trained model zero-shot on \textbf{SGD}~\citep{rastogi2020towards} and \textbf{ABCD}~\citep{chen2021abcd}; these datasets are used for transfer evaluation only, not joint training.

\subsection{Baselines}

We compare TRACER against three groups of forecasting baselines, plus recovery baselines described in Appendix~\ref{sec:appendix-recovery}: (1) \textbf{Heuristic and classical baselines}, comprising four non-trainable heuristics, \textit{No intervention} (always predicts success), \textit{Fixed-schedule confirmation} (triggers every three turns), \textit{Slot-confidence} (fires when oscillation is detected), and a \textit{feature-threshold ensemble} (predicts failure whenever any hand-designed threshold over the five trajectory features is exceeded), as well as two logistic-regression classifiers trained on trajectory features, one using only the current turn-level feature vector and one using a summarized partial-dialogue representation; (2) \textbf{Prompting baselines}, in which three instruction-tuned LLMs, Llama-3.1-8B, Mistral-7B-Instruct, and Qwen2.5-7B, are evaluated under zero-shot, zero-shot chain-of-thought (CoT), and few-shot paradigms, each receiving the serialized partial dialogue and belief state and outputting a \texttt{SUCCESS}/\texttt{FAILURE} prediction with a confidence score mapped to a continuous failure probability, with few-shot prompts including three labeled examples (one per dominant failure type) and all prompting baselines using oracle belief states (full prompt templates in Appendix~\ref{sec:appendix-prompts}); and (3) \textbf{Neural ablations}, consisting of two single-stream variants used in the ablation study (Section~\ref{sec:results-ablation}), a \textit{features-only} predictor using only the trajectory encoder and a \textit{text-only} predictor using only the belief-text encoder, which isolate the contribution of each information source before fusion.

\subsection{Metrics}

For forecasting we report AUC-ROC (ranking quality), F1 at threshold $\tau$, Early Detection Score (EDS) (normalized earliness of the first correct trigger, ranging from 1.0 at turn 1 to 0.0 if never fired), and mean detection turn $\bar{t}^*$. For threshold analysis we sweep $\tau \in [0.1, 0.9]$ and report detection rate and false positive rate (FPR); the final threshold is selected on the development set to maximize EDS subject to $\text{FPR} \leq 0.15$. Formal definitions are in Appendix~\ref{sec:appendix-metrics}.

\subsection{Implementation Details}

TRACER uses a 3-layer Transformer ($d{=}64$, 4 heads) over 5-dimensional trajectory features and a RoBERTa-base belief-text encoder with lower layers partially frozen, fused by a two-layer MLP (hidden size 256). Training uses AdamW ($\text{lr}{=}10^{-4}$, reduced encoder rate), batch size 32, and early stopping on development AUC. The trigger threshold is selected on the development split by sweeping $\tau \in [0.1, 0.9]$ to maximize EDS subject to $\text{FPR} \leq 0.15$, yielding $\tau = 0.7788$, which is then frozen for all test-time evaluation. Cross-domain evaluation is zero-shot: trained on MultiWOZ, tested on SGD and ABCD without adaptation. Full training details are in Appendix~\ref{sec:appendix-training}.

\section{Datasets and Preprocessing}
\label{sec:appendix-data}

All three datasets are converted into a common dialogue representation. Each dialogue is stored as a sequence of turns containing the user utterance, the aligned system response, a cumulative belief state (a mapping from domain-slot pairs to current values), a turn-level change set of newly modified slots, the active domain, and a dialogue-level success label. One cached split is produced per partition (train, development, test). Because the forecasting task is partial-dialogue-based, the cache stores one dialogue object per conversation; the expansion into one supervised example per visible prefix is performed at data-loading time, ensuring consistent views across all models and analyses.

\subsection{MultiWOZ 2.4}

\paragraph{Source and structure.}
MultiWOZ 2.4 is a large-scale, human-to-human, multi-domain corpus covering seven service domains: hotel, restaurant, attraction, train, taxi, hospital, and police. Every dialogue begins with a user goal specifying information requests or booking tasks, possibly spanning several domains. Raw dialogue logs are merged with processed annotation files that provide curated belief-state annotations and per-turn domain labels.

\paragraph{Turn representation.}
Each turn records: the cumulative belief state (a flat mapping of all confirmed slot values so far), the turn-level change set (slots added or revised at this turn), and the active domain from the dialogue's declared domain list.

\paragraph{Success labeling.}
For booking-capable domains (hotel, restaurant, train), a dialogue is successful only if at least one booking was confirmed. For information-only domains (attraction, taxi, hospital, police), success requires the system's final state to satisfy the user's original goal constraints. Value matching normalizes surface forms, handles pipe-separated alternatives, and allows \textit{dontcare} to match any value. Dialogues with empty goals are successful by default.

\paragraph{Data splits.}
We follow the standard train, development, and test partitions from the official MultiWOZ 2.4 release. The test partition is held out for all final reported results.

\subsection{Schema-Guided Dialogue (SGD)}

\paragraph{Source and structure.}
SGD is a large, crowd-sourced, multi-domain corpus in which each conversation involves one or more named services described by a schema. Conversations alternate strictly between user and system turns. System turns carry explicit action annotations (e.g., confirming a value, requesting a slot, notifying transaction success or failure), making it possible to infer dialogue outcome directly from the action stream.

\paragraph{Turn representation.}
For each user turn, structured frame annotations yield the active intent and cumulative slot-value pairs across all active services. Slot names are prefixed with their service identifier to prevent ambiguity. Consecutive user--system pairs are merged into unified turn objects; the belief state comes from the user turn's frame, and the change set is computed by comparing consecutive cumulative states.

\paragraph{Success labeling.}
Success is transaction-based: each dialogue is scanned for system turns carrying a success or failure notification. The final such notification determines the label. Dialogues with no transactional actions (purely informational) are treated as successful by default.

\paragraph{Data splits.}
SGD provides standard train, development, and test partitions alongside service schema definitions. The MultiWOZ-trained TRACER model is evaluated on the SGD test partition zero-shot, with no fine-tuning or schema adaptation.

\subsection{ABCD (Action-Based Conversations Dataset)}

\paragraph{Source and structure.}
ABCD is a customer-service corpus structured around the completion of operational actions rather than slot filling. Each dialogue has a scenario encoding the customer's context and a knowledge-base-defined sequence of required agent actions. Unlike MultiWOZ and SGD, ABCD provides no formal belief-state annotations; task progress is tracked through the discrete actions the agent takes.

\paragraph{Turn representation.}
Conversations involve three participant types: customer, agent, and an action track recording operational steps. Customer--agent utterances are paired into dialogue turns (de-lexicalized versions are preferred when available). Action turns are matched against a catalogue of known operational phrasings and recorded as discrete task steps.

\paragraph{Pseudo belief-state construction.}
We construct a proxy belief state by tracking when scenario values (customer name, email, order identifier, etc.) appear in utterances. Each mentioned value accumulates into a growing pseudo-state. This representation is noisier than the curated belief states in MultiWOZ and SGD, which is one reason cross-domain transfer results involving ABCD should be interpreted with caution.

\paragraph{Success labeling.}
A dialogue is successful if the agent completes at least 80\% of the knowledge-base-required actions in order. This threshold accommodates abbreviated conversations while still requiring substantial task progress.

\paragraph{Data splits.}
Train, development, and test partitions are stored in a single file with per-dialogue split annotations. The MultiWOZ-trained model is evaluated on the ABCD test partition zero-shot.

\subsection{Feature Computation}

Trajectory features are computed from the cached unified representations and stored separately per split. For MultiWOZ and SGD, the feature extractor computes the 5-dimensional trajectory vector (oscillation score, coverage rate, conflict count, fill velocity, domain shifts) at every turn. Coverage is measured against required domain slots; conflict count uses a sentence-embedding-based detector that consults the current turn delta and the previous belief state, grounding detection in actual slot changes rather than raw semantic similarity.

ABCD requires adapted proxy features because it lacks formal belief states and ontology-aligned slot tracking. Domain shifts are effectively collapsed (ABCD dialogues are single-flow), and the remaining features are interpreted through the pseudo-state and action history. The ABCD feature space is therefore comparable but not identical to that of MultiWOZ and SGD.

\subsection{Partial Dialogue Dataset Construction}

For a dialogue of length $T$, one supervised example is created for every visible partial dialogue ending at turn $t \in \{1, \dots, T\}$. Each example contains: (i) the trajectory-feature sequence from turn 1 through $t$, (ii) the serialized belief-state text for the same prefix, (iii) the final dialogue label, and (iv) the turn position and total length. This construction is shared across the full TRACER model and all ablations, ensuring all forecasting comparisons operate on the same supervision and visible context. One practical consequence is that each dialogue contributes multiple labeled examples, creating class imbalance at the partial-dialogue level; another is that the same dialogue can appear easy at some prefix lengths and hard at others, which is what enables the threshold tradeoff and fixed-context analyses.

\section{Additional Training and Implementation Details}
\label{sec:appendix-training}

\subsection{Optimization Setup}

TRACER is trained with AdamW (base learning rate $10^{-4}$; belief-text encoder at 10\% of base rate). Training uses linear warmup followed by cosine decay, with a maximum of 30 epochs, batch size 32, gradient clipping at 1.0, and early stopping (patience 5) on development-set AUC. The full model combines binary cross-entropy with the early-prediction reward ($\lambda = 0.3$). Positive-class upweighting compensates for class imbalance from partial-dialogue expansion.

\subsection{Checkpoint Selection and Threshold Freezing}

Model checkpoints are selected on the development set by development-set AUC; the same procedure applies to the features-only and text-only ablations. All final reported results use these development-selected checkpoints, reused across the threshold sweep, fixed-context evaluation, and cross-domain transfer study without retraining. The trigger threshold is then selected by sweeping $\tau \in [0.1, 0.9]$ on the development set and choosing the point that maximizes EDS subject to $\text{FPR} \leq 0.15$, yielding $\tau = 0.7788$, which is frozen for all test-time analyses. This two-stage procedure ensures that differences across experiments reflect experimental conditions rather than model configuration.

\subsection{T5 Recovery Training}

The T5-small recovery generator is fine-tuned separately from the failure predictor on MultiWOZ-derived confirmation and clarification utterances (5 epochs, batch size 16, learning rate $5\times10^{-5}$). Training inputs concatenate trajectory features, the current belief state, and recent dialogue context; targets are short recovery utterances.

\section{Additional Forecasting Results}
\label{sec:appendix-forecasting}

\subsection{Cross-Domain Transfer Results}

\begin{table}[h]
    \centering
    \small
    \setlength{\tabcolsep}{4pt}
    \begin{tabular}{lcccc}
        \hline
        \textbf{Dataset} & \textbf{AUC-ROC} & \textbf{F1} & \textbf{EDS} & \textbf{Det.\ Turn} \\
        \hline
        SGD  & 0.494 & 0.069 & 0.598 & 3.29 \\
        ABCD & 0.771 & 0.772 & 0.819 & 1.15 \\
        \hline
    \end{tabular}
    \caption{Zero-shot cross-domain transfer from a MultiWOZ-trained TRACER model. Transfer is mixed: the method remains strong on ABCD but degrades sharply on SGD.}
    \label{tab:cross-domain}
\end{table}

\subsection{Threshold Operating Points}

Table~\ref{tab:appendix-thresholds} reports all representative operating points from the threshold sweep with the full set of associated quantities. The numbers show how precision, recall, detection rate, false positive rate, and mean detection turn move together across different operating policies. Notably, the unconstrained best-EDS point (dev, $\tau{=}0.1$) achieves EDS of 1.0 only by triggering on essentially every dialogue immediately, making it an unusable intervention policy. The frozen threshold ($\tau{=}0.779$) selected under $\text{FPR} \leq 0.15$ provides a practical balance between early warning and unnecessary interventions.

\begin{table*}[t]
    \centering
    \small
    \begin{tabular}{llccccccc}
        \hline
        \textbf{Split} & \textbf{Selection} & $\tau$ & \textbf{Prec.} & \textbf{Recall} & \textbf{F1} & \textbf{EDS} & \textbf{Det. Rate} & \textbf{FPR} \\
        \hline
        dev  & Frozen threshold & 0.779 & 0.489 & 0.233 & 0.315 & 0.243 & 0.485 & 0.148 \\
        test & Frozen threshold & 0.779 & 0.552 & 0.283 & 0.374 & 0.299 & 0.540 & 0.141 \\
        dev  & Best EDS         & 0.100 & 0.199 & 1.000 & 0.331 & 1.000 & 1.000 & 1.000 \\
        dev  & Best F1          & 0.544 & 0.361 & 0.545 & 0.434 & 0.598 & 0.801 & 0.466 \\
        dev  & Lowest-FPR nontrivial & 0.900 & 0.605 & 0.106 & 0.181 & 0.118 & 0.235 & 0.041 \\
        \hline
    \end{tabular}
    \caption{Expanded threshold operating points from the MultiWOZ threshold sweep. The unconstrained best-EDS point is included for completeness but is not a practical intervention policy.}
    \label{tab:appendix-thresholds}
\end{table*}

\subsection{Fixed-Context Numeric Results}

Table~\ref{tab:appendix-context} gives the full numeric fixed-context evaluation used to support Figure~\ref{fig:partial-analysis}. The pattern is monotonic across all major metrics: as more of the dialogue becomes visible, discrimination quality, detection rate, and EDS improve steadily. This table is useful as an exact reference because the main text emphasizes the trend, while the appendix preserves the specific values associated with each visible-context condition.

\begin{table*}[t]
    \centering
    \small
    \begin{tabular}{lcccccccc}
        \hline
        \textbf{Visible Context} & \textbf{AUC-ROC} & \textbf{Precision} & \textbf{Recall} & \textbf{F1} & \textbf{Accuracy} & \textbf{EDS} & \textbf{Det. Rate} & \textbf{Mean Detect Turn} \\
        \hline
        25\%  & 0.666 & 0.510 & 0.131 & 0.209 & 0.803 & 0.103 & 0.141 & 0.61 \\
        50\%  & 0.722 & 0.500 & 0.232 & 0.317 & 0.802 & 0.159 & 0.258 & 1.63 \\
        75\%  & 0.811 & 0.566 & 0.389 & 0.461 & 0.820 & 0.240 & 0.444 & 3.08 \\
        100\% & 0.819 & 0.550 & 0.500 & 0.524 & 0.820 & 0.299 & 0.540 & 3.70 \\
        \hline
    \end{tabular}
    \caption{Numeric fixed-context forecasting results on MultiWOZ. The model recovers substantial predictive signal from partial dialogue context well before full dialogue completion.}
    \label{tab:appendix-context}
\end{table*}

\section{Triggered Recovery Analysis}
\label{sec:appendix-recovery}

\subsection{Recovery Methods}

We compare two recovery methods under the same frozen trigger policy: a \textbf{template-based module} and a \textbf{T5-based generator}. Both receive identical inputs at trigger time and differ only in how they produce a recovery utterance. Full design and implementation details are in Appendix~\ref{sec:appendix-recovery-methods}.

\subsection{Overall Recovery Comparison}

We evaluate whether TRACER's trigger policy provides a useful intervention point for lightweight recovery actions. We compare the two available recovery methods under the same frozen threshold and the same triggered dialogue set. The intervention-rate and relative-overhead columns quantify the cost of the shared trigger policy, not method-specific generation cost; they are intentionally identical across recovery methods. The central result is unambiguous: the template-based recovery strategy is substantially stronger than the learned T5 generator on our offline recovery proxy. The gap is large on groundedness (0.669 vs.\ 0.374), problem targeting (0.376 vs.\ 0.011), and clarification behavior.

This result suggests that the main challenge in the current recovery setting is not only language generation quality, but \emph{task targeting}. The T5 generator often produces short, fluent continuations, but these utterances frequently fail to request the missing information or explicitly clarify the instability that caused the trigger. By contrast, the template-based method directly leverages the detected feature pattern and current belief state, making it much more likely to produce a grounded clarification or missing-slot request. Full recovery comparison results are given in Table~\ref{tab:appendix-recovery-overall}.

\subsection{Recovery by Trigger Quality}

Table~\ref{tab:appendix-recovery-buckets} stratifies the triggered cases into \textit{Early correct}, \textit{Late correct}, and \textit{False positive} triggers. Early correct triggers test whether the system can intervene with meaningful lead time on dialogues that truly fail. Late correct triggers test whether a warning that arrives closer to the end of the dialogue still yields a useful intervention. False positives quantify the quality and cost of recovery utterances on dialogues that would have succeeded without intervention.

Template recovery remains substantially stronger than T5 across all three trigger-quality buckets. The especially large differences in \emph{problem targeting} are notable: T5 stays near zero across all buckets, indicating that its outputs often fail to address the concrete source of instability that caused the trigger.

The bucketed analysis also shows that recovery quality depends jointly on \emph{timing}, \emph{dialogue context}, and \emph{recovery method}. Taken together, these results support the conclusion that forecasting and recovery should be studied jointly, and that the current recovery setup does not yet establish a causal improvement in live end-task success.

\subsection{Expanded Overall Recovery Comparison}

Table~\ref{tab:appendix-recovery-overall} expands the overall recovery comparison by including trigger timing, utterance length, and generation-failure statistics. These additional values reinforce the interpretation in Section~6. The template-based method is consistently stronger on groundedness, problem targeting, and missing-slot requests while operating under the same intervention rate and trigger threshold. By contrast, the learned T5 generator produces shorter outputs on average and remains much weaker at explicitly targeting the instability that triggered intervention.

\begin{table*}[t]
    \centering
    \scriptsize
    \setlength{\tabcolsep}{3pt}
    \resizebox{\textwidth}{!}{%
    \begin{tabular}{lcccccccc}
        \hline
        \textbf{Recovery} & \textbf{Num} & \textbf{Grounded} & \textbf{Targeting} & \textbf{Miss-Slot} & \textbf{Clarif.} & \textbf{Mean Trigger Turn} & \textbf{Mean Trigger Fraction} & \textbf{Utterance Len.} \\
        \hline
        Template & 100 & 0.669 & 0.376 & 0.434 & 1.000 & 3.57 & 0.479 & 23.44 \\
        T5       & 100 & 0.374 & 0.011 & 0.011 & 0.570 & 3.57 & 0.479 & 14.50 \\
        \hline
    \end{tabular}%
    }
    \caption{Expanded overall recovery comparison under the frozen trigger policy. Both available recovery methods are evaluated on the same triggered dialogue set, so the differences reflect recovery quality rather than intervention frequency.}
    \label{tab:appendix-recovery-overall}
\end{table*}

\subsection{Expanded Recovery by Trigger Quality}

Table~\ref{tab:appendix-recovery-buckets} gives the fuller trigger-quality breakdown. Two patterns are worth highlighting. First, the template-based method remains stronger than T5 in every bucket, including false positives, where an unnecessary intervention can still be more or less well grounded. Second, the template outputs are longest in the late-correct bucket, which is consistent with the fact that later triggers often occur after the belief state has accumulated more information and therefore support more detailed confirmation-style recovery utterances.

\begin{table*}[t]
    \centering
    \small
    \begin{tabular}{llcccccc}
        \hline
        \textbf{Recovery} & \textbf{Trigger Quality} & \textbf{Num} & \textbf{Grounded} & \textbf{Targeting} & \textbf{Miss-Slot} & \textbf{Clarif.} & \textbf{Utterance Len.} \\
        \hline
        Template & Early correct & 19 & 0.526 & 0.307 & 0.684 & 1.000 & 18.63 \\
        Template & Late correct & 17 & 0.706 & 0.391 & 0.279 & 1.000 & 28.18 \\
        Template & False positive & 51 & 0.661 & 0.409 & 0.425 & 1.000 & 23.29 \\
        T5 & Early correct & 19 & 0.588 & 0.015 & 0.007 & 0.526 & 11.74 \\
        T5 & Late correct & 17 & 0.124 & 0.032 & 0.034 & 0.706 & 15.47 \\
        T5 & False positive & 51 & 0.362 & 0.006 & 0.007 & 0.549 & 14.14 \\
        \hline
    \end{tabular}
    \caption{Expanded recovery-by-trigger-quality results. The template method remains substantially better than T5 across all buckets, especially on problem targeting and missing-slot behavior.}
    \label{tab:appendix-recovery-buckets}
\end{table*}

\subsection{Representative Recovery Patterns}

The saved recovery outputs also make the behavioral difference between the two recovery methods more concrete. Strong template examples tend to be explicit missing-slot requests or grounded confirmation prompts that restate the currently tracked belief state and ask for the unresolved information needed to continue. Representative cases include hotel and attraction dialogues where the template recovers by naming the current domain, restating the known slots, and directly requesting the remaining fields needed to proceed. Weaker template examples are much rarer and usually occur in generic taxi clarifications such as ``are we still looking for a taxi,'' which are grammatical but weakly grounded in the actual source of instability.

The T5 generator fails in a different way. Its strongest outputs are typically fluent but underspecified, often confirming some piece of state without clearly requesting the missing information. Its weakest outputs are generic repetitions or malformed domain-shifted responses that do not address the triggering problem at all. This is consistent with the very low problem-targeting scores reported throughout the paper and helps explain why fluent generation alone is not enough for useful recovery under the shared trigger policy.

\section{Qualitative Examples}
\label{sec:appendix-qualitative}

\subsection{Forecasting Examples}

Table~\ref{tab:appendix-qualitative-examples} gives four representative examples. The false-positive examples are especially informative because they show that some so-called mistakes are semantically understandable under an early-warning framing. One representative false-positive case begins as a restaurant request, then shifts to an attraction interpretation, and later returns to restaurant constraints. TRACER raises the failure score early because the dialogue exhibits exactly the kind of domain confusion and state inconsistency that the model is designed to monitor, even though the final dataset label is success.

The false-negative cases show the opposite failure mode. In one representative case, the dialogue moves from hotel booking to attraction search to taxi planning, but the local turn-to-turn behavior remains relatively smooth. The score never crosses the trigger threshold, illustrating a failure that is hard to anticipate from partial context when the dialogue looks locally coherent until the final stages.

\begin{table*}[t]
    \centering
    \small
    \begin{tabular}{p{2.1cm}p{1.6cm}p{1.6cm}p{8.3cm}}
        \hline
        \textbf{Example Type} & \textbf{Outcome} & \textbf{Key Signal} & \textbf{Observation} \\
        \hline
        False positive & success & Domain confusion & The interaction begins as a restaurant search, shifts into attraction lookup, and later returns to restaurant constraints. TRACER fires early due to genuine cross-domain instability, even though the final outcome is success. \\
        False negative & fail & Late-revealed failure & The dialogue moves from hotel booking to attraction search to taxi planning, but local turn-to-turn behavior stays relatively smooth. The score never crosses threshold, illustrating a failure that is hard to forecast from partial context. \\
        Good recovery (template) & fail & Missing-slot targeting & The template recovery restates the known hotel name and directly requests the unresolved required slots, yielding a grounded, targeted, and actionable intervention. \\
        Bad recovery (T5) & fail & Weak recovery targeting & The T5 generator outputs a bare taxi confirmation without clarifying the unstable arrival-time field or requesting missing information, failing to address the instability that caused the trigger. \\
        \hline
    \end{tabular}
    \caption{Representative qualitative examples illustrating the main interpretation patterns: semantically understandable false positives, hard-to-anticipate false negatives, and the contrast between targeted template recovery and weak learned T5 recovery.}
    \label{tab:appendix-qualitative-examples}
\end{table*}

\subsection{Recovery Examples}

The recovery case studies further clarify what distinguishes a useful intervention from an unhelpful one. Strong template examples produce hotel or restaurant clarifications that name the currently tracked belief-state slots and explicitly request the remaining constraints needed to proceed. Even when the trigger is a false positive, the template recovery remains grounded and asks for the right missing information.

By contrast, the weak T5 examples are not merely shorter; they are misaligned with the control objective. A representative T5 failure outputs a bare taxi confirmation without addressing the unstable timing field. In another case, the generator degenerates into outputting raw feature strings (e.g., oscillation and coverage values) rather than a natural clarification move. These patterns explain why T5 underperforms so sharply on problem targeting despite producing fluent text.

\subsection{Taxonomy-Oriented Examples}

The qualitative taxonomy analysis also benefits from concrete patterns observed across sampled dialogues. Representative \textit{Contradiction / Misalignment} cases contain a stable topic but conflicting commitments or responses across turns. Representative \textit{Task Drift / Domain Confusion} cases shift away from the active task or mix incompatible domains in a way that derails progress. The noisier boundary remains between \textit{Information Incompleteness} and \textit{State Instability}: dialogues in these categories often co-exhibit missing slots and repeated revisions, which partly explains why those two categories retain the small amount of disagreement reported in Table~\ref{tab:taxonomy-summary}.

\section{Annotation Details}
\label{sec:appendix-taxonomy}

\subsection{Taxonomy Definitions}

Table~\ref{tab:taxonomy-definitions} provides the formal definitions of the four failure categories used in the qualitative annotation study.

\begin{table}[ht]
    \centering
    \small
    \begin{tabular}{p{2.2cm}p{5.0cm}}
        \toprule
        \textbf{Category} & \textbf{Definition} \\
        \midrule
        Information Incompleteness & The task fails because required constraints, booking attributes, or other task-critical details are never stably assembled. The dialogue ends without the user's needs fully specified. \\
        \midrule
        Contradiction / Misalignment & The system and user are misaligned on facts or constraints, leading to explicit corrections, unresolved contradictions, or repeated repair sequences. \\
        \midrule
        State Instability & Slot values or the dialogue state flip repeatedly without converging to a stable representation. The belief state oscillates without reaching a coherent final state. \\
        \midrule
        Task Drift / Domain Confusion & The interaction shifts away from the active task or mixes incompatible domains, causing the system to respond to the wrong goal or lose track of the user's current objective. \\
        \bottomrule
    \end{tabular}
    \caption{Definitions of the four failure categories in the TRACER qualitative taxonomy. These categories are used as an interpretive analysis framework and are not treated as ground-truth supervision.}
    \label{tab:taxonomy-definitions}
\end{table}

\subsection{Sampling and Annotation Setup}

The annotation study reported in the paper is based on a balanced sample of 60 failed MultiWOZ dialogues. We first assign each dialogue a heuristic primary label using the saved failure-analysis pipeline, then sample 15 dialogues from each heuristic category. This produces a balanced analysis set with 15 examples each for Information Incompleteness, Contradiction / Misalignment, State Instability, and Task Drift / Domain Confusion.

The sampled dialogues are then passed through a human annotation step. The purpose of this stage is not to create ground-truth labels for model training. Instead, it provides a structured second-pass interpretation of the sampled failures so that we can assess whether the taxonomy tracks recognizable failure patterns. The resulting annotations are therefore used only as qualitative evidence in support of the analysis sections of the paper.

\subsection{Observed Label Distribution}

Within the 60-dialogue sample, the heuristic labeling is exactly balanced by construction, with 15 dialogues per class. The human-annotated primary-label distribution is slightly different: 18 dialogues are assigned to Contradiction / Misalignment, 16 to Task Drift / Domain Confusion, 13 to Information Incompleteness, and 13 to State Instability. This small redistribution is consistent with the main-paper result that contradiction and drift are comparatively easy to recognize, while incompleteness and instability remain somewhat harder to separate cleanly.

\subsection{Heuristic--Human Alignment Pattern}

The main alignment pattern can be summarized as follows. All 15 dialogues heuristically labeled as Contradiction / Misalignment remain on-diagonal under the human annotation, and the same is true for all 15 Task Drift / Domain Confusion cases. Information Incompleteness keeps 13 of 15 cases on-diagonal, with one case moving to Contradiction / Misalignment and one to Task Drift / Domain Confusion. State Instability also keeps 13 of 15 cases on-diagonal, with the remaining two cases moving to Contradiction / Misalignment.

This pattern is important for interpretation. It suggests that the taxonomy is quite stable for contradiction-heavy and task-drift-heavy failures, but still retains a soft boundary between incompleteness and instability. That is a sensible qualitative outcome: missing information and unstable state updates often co-occur in practice, especially in longer task-oriented dialogues where slot filling and slot revision interact over time.

\subsection{Scope and Limitations of the Annotation Study}

This appendix should be read with the same caveat stated in the main paper: these annotations are not gold labels. The qualitative study is intended to help interpret what kinds of failures TRACER tends to detect, not to define a new benchmark for supervised taxonomy prediction. The balanced sampling design is useful for analysis because it ensures coverage of all four categories, but it should not be mistaken for the true label distribution of failures in the underlying dataset. Likewise, the human annotation adds a structured second opinion, but it does not replace further adjudication or interactive user studies.

\section{Evaluation Metric Definitions}
\label{sec:appendix-metrics}

This appendix provides formal mathematical definitions for all evaluation metrics used in the paper. Let $\hat{p}_{i,t} \in [0,1]$ denote the predicted failure probability for dialogue $i$ at turn $t$, $y_i \in \{0,1\}$ the terminal label ($y_i=1$ for failure), $\mathcal{F} = \{i: y_i = 1\}$ the set of failed dialogues, $\mathcal{S} = \{i: y_i = 0\}$ the set of successful dialogues, and $T_i$ the total length of dialogue $i$. A threshold $\tau$ converts continuous scores into binary predictions $\hat{y}_{i,t} = \mathbf{1}[\hat{p}_{i,t} \geq \tau]$.

\subsection{Forecasting Metrics}

\paragraph{AUC-ROC.}
The area under the receiver operating characteristic curve equals the probability that the model ranks a random failing instance above a random succeeding one across all partial-dialogue pairs:
\begin{equation*}
\text{AUC-ROC} = \mathbb{P}\bigl(\hat{p}_{i,t} > \hat{p}_{j,t'} \mid y_i = 1,\; y_j = 0\bigr)
\label{eq:auc-app}
\end{equation*}

\paragraph{Early Detection Score (EDS).}
Let $t_i^* = \min\{t : \hat{p}_{i,t} > \tau\}$ be the first turn at which the trigger fires for dialogue $i$ ($t_i^* = \infty$ if it never fires). The per-dialogue EDS is the normalized detection earliness:
\begin{equation*}
\text{EDS}_i = \frac{T_i - t_i^*}{T_i}
\label{eq:eds-app}
\end{equation*}
$\text{EDS}_i = 1$ when detection occurs at the first turn; $\text{EDS}_i = 0$ when no trigger fires. The corpus-level EDS averages over all failed dialogues:
\begin{equation*}
\text{EDS} = \frac{1}{|\mathcal{F}|} \sum_{i \in \mathcal{F}} \text{EDS}_i
\label{eq:eds-corpus}
\end{equation*}

\paragraph{Mean Detection Turn.}
\begin{equation*}
\bar{t}^* = \frac{1}{|\{i \in \mathcal{F}: t_i^* < \infty\}|} \sum_{\substack{i \in \mathcal{F} \\ t_i^* < \infty}} t_i^*
\label{eq:mean-det-turn}
\end{equation*}

\subsection{Threshold Tradeoff Metrics}

\paragraph{Detection Rate and False Positive Rate.}
\begin{equation*}
\text{DetRate} = \frac{|\{i \in \mathcal{F} : t_i^* < \infty\}|}{|\mathcal{F}|}
\label{eq:det-rate}
\end{equation*}

\begin{equation*}
\text{FPR} = \frac{|\{i \in \mathcal{S} : \exists\, t,\; \hat{p}_{i,t} > \tau\}|}{|\mathcal{S}|}
\label{eq:fpr}
\end{equation*}
A dialogue in $\mathcal{S}$ contributes to the FPR numerator if the trigger fires at \emph{any} turn, not just at one specific turn. This definition penalizes policies that fire frequently on successful dialogues regardless of when they fire.

\subsection{Recovery Quality Metrics}

For each triggered dialogue, TRACER fires at the first turn $t^*$ where $\hat{p}_{i,t^*} > \tau$. A recovery utterance $r$ is then generated and scored against the dialogue state at turn $t^*$.

\paragraph{Grounded ($G$).}
Measures whether $r$ is grounded in the current belief state $b_{t^*}$. Up to 6 non-empty belief-state slot-value pairs are considered. For each slot $s$ with value $v$, a match occurs if either the slot name (in any of its surface variants) or the slot value appears in the normalized text of $r$:
\begin{equation*}
\small
G(r) = \tfrac{\bigl|\{(s,v) \in b_{t^*}^{(6)} : \text{slot\_match}(s,r) \vee \text{value\_match}(v,r)\}\bigr|}{\bigl|\,b_{t^*}^{(6)}\,\bigr|}
\label{eq:grounded}
\end{equation*}
where $b_{t^*}^{(6)}$ denotes the first 6 non-empty slots of $b_{t^*}$ sorted lexicographically.

\paragraph{Targeting ($P$).}
Measures whether $r$ addresses the \emph{problem slots}: slots that are missing from the current belief state (required but unfilled) or that become unstable after the trigger turn (their value changes in at least one future turn). Let $\mathcal{M}$ be the set of missing required slots and $\mathcal{U}$ the set of future-unstable slots. The combined problem set is $\Pi = \mathcal{M} \cup \{s\text{'s local name}: s \in \mathcal{U}\}$:
\begin{equation*}
P(r) = \frac{|\{s \in \Pi : \text{slot\_match}(s, r)\}|}{|\Pi|}
\label{eq:targeting}
\end{equation*}
If $\Pi = \emptyset$, then $P(r) = 0$.

\paragraph{Missing-Slot Request ($M$).}
Measures whether $r$ explicitly requests the \emph{missing required} slots only (not unstable slots):
\begin{equation*}
M(r) = \frac{|\{s \in \mathcal{M} : \text{slot\_match}(s, r)\}|}{|\mathcal{M}|}
\label{eq:missslot}
\end{equation*}
If $\mathcal{M} = \emptyset$, then $M(r) = 0$.

\paragraph{Clarification Intent ($C$).}
A binary indicator that checks whether $r$ contains a question mark or any of the following keywords: \textit{confirm, clarify, correct, right, make sure, just to, could you provide, could you confirm, is that, would you like, i still need}:
\begin{equation*}
\small
C(r) = \mathbf{1}\bigl[\text{``?''} \in r \;\vee\; \exists\, k \in \mathcal{K},\; k \in \text{lower}(r)\bigr]
\label{eq:clarif}
\end{equation*}
where $\mathcal{K}$ is the keyword set above.

\paragraph{Intervention Rate.}
The fraction of test dialogues for which the trigger fires at least once:
\begin{equation*}
\text{Interv. Rate} = \frac{|\{i \in \mathcal{D}_{\text{test}} : t_i^* < \infty\}|}{|\mathcal{D}_{\text{test}}|}
\label{eq:interv-rate}
\end{equation*}

\paragraph{Relative Overhead.}
Under the assumption that each recovery intervention adds one extra dialogue turn, the relative overhead is the intervention rate normalized by the mean dialogue length $\bar{T}$:
\begin{equation*}
\text{Rel. Overhead} = \frac{\text{Interv. Rate}}{\bar{T}}
\label{eq:rel-overhead}
\end{equation*}

\begin{equation*}
\bar{T} = \frac{1}{|\mathcal{D}_{\text{test}}|}\sum_{i \in \mathcal{D}_{\text{test}}} T_i
\label{eq:avg-turns}
\end{equation*}
Both Template and T5 recovery share the same Interv. Rate and Rel. Overhead because they operate under the same TRACER trigger policy; only the generated utterance differs.

\section{Recovery Method Details}
\label{sec:appendix-recovery-methods}

\subsection{Triggered Recovery Policy}

The predictor becomes a control layer by thresholding the partial-dialogue-level failure score. For a partial dialogue at turn $t$, if $p(y=1 \mid D_{\leq t}) > \tau$, the system triggers a recovery action. We select $\tau$ on the development set under a false-positive-rate constraint and then freeze that operating point for test-time evaluation. This distinction matters because unconstrained low thresholds can maximize early-detection metrics trivially while producing impractically many unnecessary interventions.

We study two recovery methods under this shared trigger policy. The first is a \textbf{template-based recovery module}, which maps dominant feature patterns such as oscillation, coverage gaps, contradiction, or domain confusion to short clarification or confirmation prompts. The second is a \textbf{T5-based recovery generator}, implemented with a fine-tuned T5 model~\citep{raffel2020exploring}. Its input concatenates the current trajectory features, the current belief state, and the most recent dialogue context, and it generates a short recovery utterance intended to confirm the user goal or request missing information.

This appendix describes the two recovery methods in detail. Both methods receive the same input at trigger time: the trajectory feature vector $\mathbf{f}_{t^*}$, the current belief state $b_{t^*}$, the active domain, and a short window of recent dialogue turns. They differ only in how they map this input to a recovery utterance.

\subsection{Template-Based Recovery}

The template-based recovery module first classifies the dominant failure type from the trigger-turn feature vector, then selects a template from a small fixed set and fills it with the current belief-state content.

\paragraph{Failure type detection.}
The dominant failure type is determined by thresholding the trajectory features in priority order:
\begin{enumerate}
    \item \textbf{Oscillation} ($\text{OscScore}_{t^*} \geq 2$): the belief state is oscillating; use a confirmation template.
    \item \textbf{Coverage gap} ($\text{CoverageRate}_{t^*} < 0.4$): required slots are missing; use a missing-slot request template.
    \item \textbf{Conflict} ($\text{ConflictCount}_{t^*} \geq 2$): repeated contradictions detected; use a clarification template.
    \item \textbf{Domain confusion} ($\text{DomainShifts}_{t^*} \geq 3$): excessive domain switching; use a domain-focus template.
    \item \textbf{General} (none of the above): use a general confirmation template.
\end{enumerate}

\paragraph{Template structure.}
Each failure type maps to 2--3 surface-form templates. For example, the coverage-gap type maps to utterances of the form:

\begin{quote}
\textit{``To complete your \{domain\} request, I still need: \{missing\_slots\}. Could you provide those?''}
\end{quote}

\noindent and the oscillation type maps to utterances of the form:

\begin{quote}
\textit{``I notice some changes in your preferences. Just to confirm, you'd like a \{domain\} with \{slot\_text\}?''}
\end{quote}

One template per type is selected uniformly at random and filled with the domain name, the formatted belief-state slots, and the list of missing required slots. The resulting utterance is always a short, directly actionable clarification or confirmation move.

\subsection{T5-Based Recovery}

The learned recovery model is a fine-tuned T5-small sequence-to-sequence generator~\citep{raffel2020exploring}. It is trained separately from the failure predictor on MultiWOZ-derived confirmation and clarification utterances from the training split.

\paragraph{Input format.}
Each training and inference input concatenates three information sources in a fixed order:
\begin{enumerate}
    \item A string rendering of the five trajectory features at the trigger turn, e.g.\ \texttt{osc=1.0 cov=0.33 conflict=1 velocity=0.5 domains=2}.
    \item A compact serialization of the current belief state, e.g.\ \texttt{hotel-area: west, hotel-pricerange: cheap}.
    \item The most recent 3 dialogue turns formatted as alternating \texttt{USER:} and \texttt{SYSTEM:} lines.
\end{enumerate}

\paragraph{Training.}
The T5-small generator is fine-tuned for 5 epochs with batch size 16 and learning rate $5 \times 10^{-5}$. Training targets are system utterances from the training split that function as confirmations or recovery moves (e.g.\ utterances following a belief-state correction or preceding a booking action). Training uses a linear warmup schedule over 10\% of total steps.

\paragraph{Inference.}
At trigger time, the input is assembled from the live trajectory features, belief state, and recent context. Generation uses greedy decoding with a maximum of 128 new tokens. The T5 checkpoint is loaded once and reused for all triggered dialogues in the evaluation set.

As shown in the main paper, the T5 generator substantially underperforms the template method on problem targeting despite producing fluent text. This outcome highlights that the challenge in recovery is not language generation quality but correct identification and addressing of the instability source, a property that the template method handles directly through its feature-driven routing logic.

\section{LLM Prompt Templates}
\label{sec:appendix-prompts}

This appendix provides the prompt templates used in two distinct LLM inference pipelines: (1) zero-shot failure prediction, where an instruction-tuned model classifies each partial dialogue as likely to succeed or fail; and (2) generated belief-state tracking, where a model predicts cumulative belief states turn by turn.

\subsection{Zero-Shot Failure Prediction Prompts}
\label{sec:prompt-failure}

These prompts are used to evaluate Llama-3.1-8B, Mistral-7B-Instruct, and Qwen2.5-7B as zero-shot failure forecasters (the \textit{Zero-Shot Prompting} rows in Table~\ref{tab:main-forecasting}). Each model receives the same system and user prompts.

\begin{tcolorbox}[colback=gray!5!white, colframe=black, fontupper=\small, width=\linewidth, boxsep=0pt, title=System Prompt]You are evaluating a task-oriented dialogue prefix. Your job is to forecast whether the full dialogue will eventually end in SUCCESS or FAILURE. Use only the visible prefix. Respond with valid JSON only.\end{tcolorbox}

\paragraph{User prompt template.}
The following template is instantiated for every partial dialogue $(i,t)$. \texttt{\{dialogue\_block\}} is the turn-by-turn prefix with \texttt{Turn $k$ USER:} / \texttt{Turn $k$ SYSTEM:} prefixes; \texttt{\{belief\_text\}} is the oracle cumulative belief state serialized as \texttt{"domain-slot: value"} pairs.

\begin{tcolorbox}[colback=gray!5!white, colframe=black, fontupper=\small, width=\linewidth, boxsep=0pt, title=User Prompt]Given the task-oriented dialogue prefix below, predict whether the full dialogue will eventually succeed or fail.\\[4pt] Rules:\begin{itemize} \item Predict FAILURE if the dialogue appears likely to end without satisfying the user's full goal. \item Predict SUCCESS if the visible evidence suggests the task is likely to be completed. \item Use the visible prefix only. Do not assume future repair unless strongly supported. \item Return valid JSON only. \end{itemize} Visible prefix:\\ \{dialogue\_block\}\\[4pt] Current cumulative belief state:\\ \{belief\_text\}\\[4pt] Return JSON with exactly this schema:\\ \{ \\ \hspace{1em}``prediction'': ``FAILURE'' or ``SUCCESS'',\\ \hspace{1em}``confidence'': $<$integer from 0 to 100$>$,\\ \hspace{1em}``rationale'': ``$<$one short sentence$>$''\\ \}\end{tcolorbox}

The confidence score is mapped to a failure probability via $\hat{p} = \texttt{confidence}/100$ when \texttt{prediction} is \texttt{"FAILURE"} and $\hat{p} = 1 - \texttt{confidence}/100$ otherwise, yielding a continuous score for AUC computation.

\subsection{Generated Belief-State Prompts}
\label{sec:prompt-belief}

These prompts are used to generate predicted belief states with Qwen3-30B-A3B. The predicted belief states replace oracle states in the \textit{TRACER (generated)} row of Table~\ref{tab:main-forecasting}. The pipeline processes all dialogues turn by turn in lockstep, feeding each model output as the \texttt{previous\_belief\_state} for the next turn.

\paragraph{System prompt (\texttt{constrained\_careful}).}
This is the primary system prompt used for belief-state generation. It enforces the MultiWOZ domain--slot ontology and provides detailed tracking rules.

\begin{tcolorbox}[colback=gray!5!white, colframe=black, fontupper=\small, width=\linewidth, boxsep=0pt, title=System Prompt]You are a precise dialogue state tracker for MultiWOZ-style task-oriented dialogue. Predict the cumulative belief state after the latest USER utterance.\\[4pt] Rules:\begin{itemize} \item Use only the dialogue text shown in the prompt. \item Output only valid JSON. Do not include markdown, explanations, or comments. \item The JSON must have exactly these top-level keys: ``domain'' and ``belief\_state''. \item ``domain'' must be the active domain/service after the latest USER utterance. \item ``belief\_state'' must be a JSON object mapping slot keys to string values. \item Slot keys must use the format ``domain-slot'', for example ``restaurant-area''. \item Use lowercase domain and slot names. \item Start from the previous belief\_state given in the prompt. \item Return the full cumulative belief\_state, not only the new changes. \item Keep earlier slot values across domains unless the user explicitly corrects, replaces, or cancels them. \item If the user corrects a domain, move the value to the corrected domain and remove the wrong-domain version. \item Track user constraints such as area, food, price range, type, name, day, time, people, departure, destination, leave time, and arrival time. \item Track booking information when the user gives booking day, time, stay length, or number of people. \item If the system offers a specific entity and the user accepts it or asks follow-up details about it, include that entity as the corresponding ``domain-name''. \item If the user asks for address, postcode, phone, entrance fee, reference number, rating, or car type, treat these as requested information, not belief-state slots. \item If the user says ``same area'', copy the relevant earlier area value to the new domain. \item Do not hallucinate values from database results unless the dialogue text makes the user accept or refer to that entity. \item Omit unknown, not mentioned, none, empty, or unsupported slots. \item Values should be short normalized strings, not full sentences. \end{itemize}\end{tcolorbox}

\paragraph{User prompt template (\texttt{constrained} style).}
The following template is instantiated at each turn $t$. \texttt{\{schema\}} lists allowed \texttt{domain: slot, slot, ...} lines; \texttt{\{prefix\}} is the dialogue context rendered as \texttt{Turn $k$ USER:} / \texttt{Turn $k$ SYSTEM:} lines; and \texttt{\{previous\_json\}} is the JSON-serialized belief state from the previous turn.

\begin{tcolorbox}[colback=gray!5!white, colframe=black, fontupper=\small, width=\linewidth, boxsep=0pt, title=User Prompt]Allowed domains and slots:\\ \{schema\}\\[4pt] Dialogue prefix:\\ \{prefix\}\\[4pt] Previous belief\_state before the latest USER utterance:\\ \{previous\_json\}\\[4pt] Return the belief state after the latest USER utterance.\\ Required JSON format:\\ \{"domain": "restaurant", "belief\_state": \{"restaurant-area": "north"\}\}\end{tcolorbox}

Slot alias normalization is applied to model outputs before the belief state is stored: for example, \texttt{centre}/\texttt{center}~$\to$~\texttt{area}, \texttt{price}~$\to$~\texttt{pricerange}, \texttt{depart}~$\to$~\texttt{leaveat}. Values are lowercased and the special form \texttt{"center"} is normalized to \texttt{"centre"}. If the model output cannot be parsed as valid JSON, a regex fallback extracts the innermost JSON object from the raw output string. The turn delta $\Delta_t$ is then computed by comparing the predicted belief state against the previous turn's predicted state.

\end{document}